\documentclass{article}

\usepackage[table]{xcolor}

\usepackage[final]{corl_2026} % Uncomment for the camera-ready ``final'' version.

\usepackage{amssymb}
\usepackage{multirow}
\usepackage{bigdelim}
\usepackage{longtable}
\usepackage{tabularray}
\usepackage{wrapfig}
\usepackage[most]{tcolorbox} 
\usepackage{url}
\usepackage{xspace}
\usepackage{svg}
\usepackage[absolute]{textpos} %
\usepackage{fdsymbol}   %
\usepackage{csvsimple}
\usepackage[utf8]{inputenc} %
\usepackage[T1]{fontenc}    %
\usepackage{url}            %
\usepackage{booktabs}       %
\usepackage{amsfonts}       %
\usepackage{nicefrac}       %
\usepackage{microtype}      %
\usepackage{amsmath}
\usepackage{csquotes}
\usepackage{siunitx}
\usepackage{graphicx}
\usepackage{arydshln}
\usepackage{wrapfig}
\usepackage{enumitem}
\usepackage{soul} %
\usepackage{multirow}
\usepackage{xspace}
\usepackage{adjustbox}
\usepackage{pifont}
\usepackage[font=small, labelfont=bf]{caption}
\usepackage{makecell}
\usepackage{bold-extra}
\usepackage{url}
\usepackage{booktabs,rotating}
\usepackage{setspace}
\usepackage{quoting}      % nice quote environment
\usepackage{float}
\usepackage{nicematrix}
\usepackage{hyperref}
\usepackage{listings}
\usepackage{tikz}
\usepackage{textcomp}
\usepackage{adjustbox}
\usepackage{placeins}
\usepackage{stfloats}

\usepackage{array}
\newcolumntype{P}[1]{>{\centering\arraybackslash}p{#1}}

\makeatletter
\renewcommand\subsection{\@startsection{subsection}{2}{\z@}%
                         {3pt \@plus 1ex \@minus .2ex}%   <-- before
                         {2pt \@plus .2ex}%               <-- after
                         {\normalfont\bfseries}}
\makeatother

\definecolor{ai2pink}{HTML}{f0529c}
\definecolor{ai2midpink}{HTML}{fad3e5}
\definecolor{ai2lightpink}{HTML}{fbecf3}
\definecolor{ai2midwhite}{HTML}{f2e5d9}
\definecolor{ai2offwhite}{HTML}{fbf4ee}
\definecolor{ai2green}{HTML}{0fcb8c}
\definecolor{ai2lightgreen}{HTML}{e7f9f3}
\definecolor{ai2darkgreen}{HTML}{105257}
\definecolor{ai2purple}{HTML}{B932EB}
\definecolor{ai2lightpurple}{HTML}{f7e8fc}

\newcommand{\pizero}{$\pi_0\text{-DROID}$}
\newcommand{\pizerofive}{$\pi_{0.5}\text{-DROID}$}
\newcommand{\pizerofiveft}{$\pi_{0.5}\text{-DROID-Finetune}$}

\newcommand{\modelfamily}{MolmoBot\xspace}
\newcommand{\molmomodel}{MolmoBot\xspace}
\newcommand{\molmomodelimg}{\molmomodel-Img\xspace}

\newcommand{\molmomodelmultiframe}[1]{\molmomodel (H=#1)\xspace}

\newcommand{\trainingdata}{MolmoBot-Data\xspace}
\newcommand{\trainingdataengine}{MolmoBot-Engine\xspace}
\newcommand{\spocmodel}{MolmoBot-SPOC\xspace}
\newcommand{\paligemmamodel}{MolmoBot-Pi0\xspace}

\newcommand{\molmoshort}{MB\xspace}
\newcommand{\spocshort}{MB-SPOC\xspace}

\newcommand{\myparagraph}[1]{\textbf{#1}}

\newcommand{\cmark}{\ding{51}}%
\newcommand{\xmark}{\ding{55}}%

\newif\ifshowcomments
\showcommentstrue
\ifshowcomments
\newcommand{\ranjay}[1]{{\color{orange} $[$#1$]^R_K$}}
\newcommand{\dieter}[1]{{\color{violet} [dieter]: #1}}
\newcommand{\maxa}[1]{{\color{darkgreen} $[$#1$]^M_A$}}
\newcommand{\rmh}[1]{{\color{olive} [Rose]: #1}}
\newcommand{\omar}[1]{{\color{green} [Omar]: #1}}
\newcommand{\yejink}[1]{{\color{red} [Yejin]: #1}}
\newcommand{\winson}[1]{{\color{cyan} [Winson]: #1}}
\newcommand{\jordis}[1]{{\color{brown} [Jordi: #1]}}
\newcommand{\abhayd}[1]{{\color{blue} [Abhay: #1]}}
\newcommand{\mayag}[1]{{\color{teal} [Maya: #1]}}
\newcommand{\arjung}[1]{{\color{magenta} [Arjun: #1]}}
\newcommand{\snehalj}[1]{{\color{darkred} [Snehal: #1]}}
\newcommand{\wilbertp}[1]{{\color{violet} [Wilbert: #1]}}
\newcommand{\roseh}[1]{{\color{lime} [Rose: #1]}}
\newcommand{\ainaz}[1]{{\color{brown} [Ainaz: #1]}}

\newcommand{\rohun}[1]{{\color{pink} [Rohun: #1]}}
\newcommand{\dhruv}[1]{{\color{orange} [Dhruv: #1]}}
\else
  \newcommand{\ranjay}[1]{}
  \newcommand{\dieter}[1]{}
  \newcommand{\maxa}[1]{}
  \newcommand{\rmh}[1]{}
  \newcommand{\omar}[1]{}
  \newcommand{\MS}[1]{}
  \newcommand{\yejink}[1]{}
  \newcommand{\winson}[1]{}
  \newcommand{\jordis}[1]{}
  \newcommand{\abhayd}[1]{}
  \newcommand{\mayag}[1]{}
  \newcommand{\arjung}[1]{}
  \newcommand{\snehalj}[1]{}
  \newcommand{\wilbertp}[1]{}
  \newcommand{\roseh}[1]{}
  \newcommand{\ainaz}[1]{}
  \newcommand{\rohun}[1]{}
  \newcommand{\dhruv}[1]{}
  \newcommand{\todo}[1]{}
\fi

\title{MolmoB0T: Large-Scale Simulation \\ Enables Zero-Shot Manipulation}

\newcommand{\core}{\textsuperscript{\textcolor{ai2pink}{\ding{170}}}}
\newcommand{\coremark}{\textcolor{ai2pink}{\ding{170}}}
\newcommand{\coadvise}{\textsuperscript{\textdagger}}

\author{
{\normalfont
Abhay Deshpande\core\textsuperscript{1*},
Maya Guru\core\textsuperscript{1*},
Rose Hendrix\core\textsuperscript{1*},
Snehal Jauhri\core\textsuperscript{1,4*},
}\\
Ainaz Eftekhar\core\textsuperscript{1,2},
Rohun Tripathi\core\textsuperscript{1},
Max Argus\core\textsuperscript{1},
Jordi Salvador\core\textsuperscript{1}, \\
Haoquan Fang\core\textsuperscript{1,2},
Matthew Wallingford\core\textsuperscript{1},
Wilbert Pumacay\core\textsuperscript{1},
Yejin Kim\core\textsuperscript{1}, \\
Quinn Pfeifer\textsuperscript{2},
Ying-Chun Lee\textsuperscript{2},
Piper Wolters\textsuperscript{1},
Omar Rayyan\textsuperscript{3},
Mingtong Zhang\textsuperscript{5}, \\
Jiafei Duan\textsuperscript{1,2},
Karen Farley\textsuperscript{1},
Winson Han\textsuperscript{1},
Eli Vanderbilt\textsuperscript{1}, \\
Dieter Fox\textsuperscript{1,2},
Ali Farhadi\textsuperscript{1,2},
Georgia Chalvatzaki\textsuperscript{4}, \\
Dhruv Shah\core\coadvise\textsuperscript{5},
Ranjay Krishna\core\coadvise\textsuperscript{1,2} \\
\textsuperscript{1}Allen Institute for AI, \textsuperscript{2}University of Washington, \textsuperscript{3}University of California, Los Angeles, \\
\textsuperscript{4}Technische Universität Darmstadt, \textsuperscript{5}Princeton University \\
*denotes equal contribution in alphabetical order, \\
\coremark{} marks core contributors, and \textdagger{} indicates equal advising.
}

\begin{document}
\maketitle

%===============================================================================

\begin{figure}[h!]
  \centering
  \vspace{-0.75cm}
  \includegraphics[width=\linewidth]{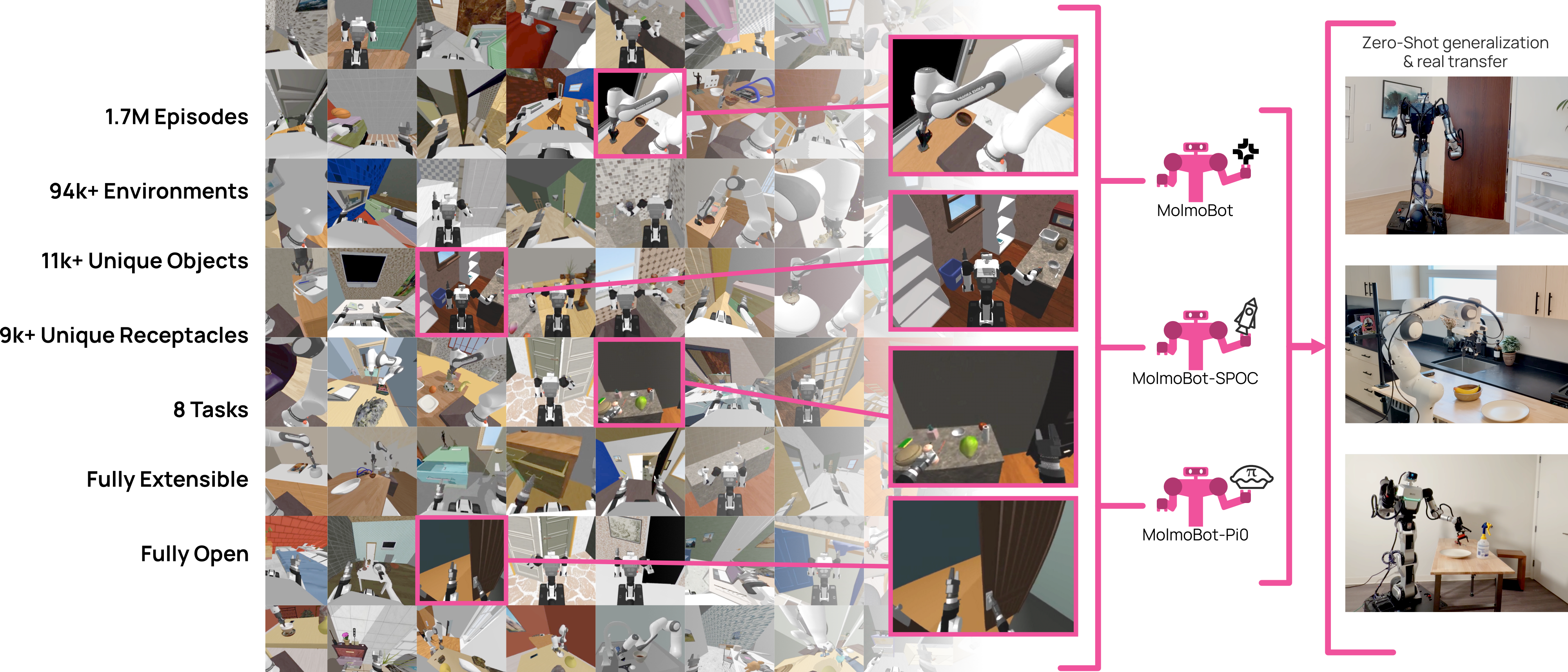}
  \caption{\textbf{\molmomodel{}} leverages diverse simulation data to achieve zero-shot sim-to-real transfer on multiple robotic tasks such as pick-and-place and door opening. This unlocks the ability to dramatically scale up the training data for generalist robotic foundation models.}
  \label{fig:example}
\end{figure}

\vspace{-0.2cm}
\begin{abstract}
A prevailing view in training generalist policies is that simulation alone is not enough; effective sim-to-real transfer is widely believed to require at least some real-world data collection or task-specific fine-tuning to bridge the gap between simulated and physical environments. In this paper, we challenge that assumption: 
with sufficiently large-scale and diverse synthetic training data, we show that zero-shot transfer to the real world is not only possible, but state-of-the-art for tabletop manipulation, with demonstrated extension to complex articulated and mobile manipulation tasks.
We introduce \emph{\mbox{\trainingdataengine{}}}, a fully open-source pipeline for procedural data generation across robots, tasks, and diverse simulated environments in MolmoSpaces. With it, we release \emph{\trainingdata{}}, a dataset of 1.7 million expert trajectories for articulated object manipulation and pick-and-place tasks. 
We train three policy classes: \emph{\molmomodel{}}, a Molmo2-based multi-frame vision-language model with a flow-matching action head; \emph{\paligemmamodel{}}, which replicates the $\pi_0$ architecture to enable direct comparison; and \emph{\spocmodel{}}, a lightweight policy suitable for edge deployment and amenable to RL fine-tuning. 
We evaluate on two robotic platforms: the Franka FR3 for tabletop manipulation tasks and the Rainbow Robotics RB-Y1 mobile manipulator for door opening, drawer manipulation, cabinet interaction, and mobile pick-and-place. 
Without any real-world fine-tuning, our policies achieve zero-shot sim-to-real transfer with unseen objects and environments for tabletop and door-opening tasks. On tabletop pick-and-place, \molmomodel{} achieves a success rate of 79.2\% in real world evaluations across 4 settings, outperforming $\pi_{0.5}$ at 39.2\%. Our results demonstrate that procedural environment generation combined with diverse articulated assets can produce robust manipulation policies that generalize broadly to the real world. For videos and additional information, please see \href{https://allenai.github.io/MolmoBot/}{our project website}.
\end{abstract}

% \keywords{Robotics, Manipulation, Simulation, Sim2Real} 

\section{Introduction}

Recent work in robotic manipulation \cite{black2024pi_0,black2025pi_05,nvidia2025gr00tn1openfoundation,team2025gemini} has made impressive progress towards robotics foundation models, producing robust manipulation policies increasingly capable of environment and task generalization. A key factor that drives these advances is large-scale real-world data sourced from expert tele-operation. These datasets are very expensive to collect, both in terms of money and human effort, and are therefore difficult to continue to scale. Generating synthetic data via simulation at scale could provide a salve to this issue, owing to its cheap scalability and high controllability, but widespread adoption is presently limited --- current works largely only use synthetic data for narrow-domain training~\cite{yin2026emergent,xue2025doorman} or as a small portion of the pre-training mixture before additional post-training with real data~\cite{nvidia2025gr00tn1openfoundation,tian2025interndata}. This limited adoption is typically rooted in the domain gap in physical and visual realism of today's physics-based simulation engines.

This domain gap is quite challenging to overcome, traditionally requiring either painstaking system identification~\cite{memmel2024asid} or techniques such as iterative adaptation~\cite{ren2023adaptsim} or co-training~\cite{maddukuri2025simandreal}. In this paper, however, we find that this gap can be mitigated by leveraging massive scale and extreme diversity. We conclusively demonstrate that synthetic data generated via procedural simulation is not only sufficient for building zero-shot generalists, but also effective at training state-of-the-art generalists.

We introduce MolmoBot-Engine, a data generation pipeline with many robots, tasks, and diverse simulated environments, and MolmoBot-Data, a dataset of 1.7M demonstrations spanning articulated and rigid object manipulation across multiple robots. We train three policy classes: MolmoBot, our flagship VLA, MolmoBot-Pi0, a direct comparison to $\pi_0$, and MolmoBot-SPOC, a lightweight policy for edge deployment. Evaluated on the Franka FR3 and Rainbow Robotics RB-Y1 without any real-world fine-tuning, our policies transfer zero-shot to unseen objects and environments. MolmoBot achieves 79.2\% success on tabletop pick-and-place versus 39.2\% for $\pi_{0.5}$, and MolmoBot-Pi0 reaches 46.5\% using the same architecture, demonstrating that the performance gap is driven by data. Our results suggest the barrier to general-purpose sim-to-real manipulation is less about an irreducible domain gap and more about access to sufficiently large, diverse, and open simulation pipelines; we provide that access by open-sourcing all components.

\section{Related Work}
\label{sec:related}
% \myparagraph{Imitation learning for manipulation.}
Imitation learning has emerged as a dominant paradigm for robot manipulation, with methods ranging from behavior cloning~\cite{pomerleau2015alvinn,Zhang2017DeepIL} to more sophisticated approaches incorporating temporal abstractions~\cite{Lynch2019LearningLP} or diffusion models~\cite{Chi2023DiffusionPV}. RT-1~\cite{Brohan2022RT1RT} and RT-2~\cite{Brohan2023RT2VM} demonstrated that scaling model capacity and multi-task data produces capable manipulation policies, which has been extended by multiple policy families --- including $\pi$~\cite{black2024pi_0,black2025pi_05}, GR00T~\cite{nvidia2025gr00tn1openfoundation}, and Gemini Robotics~\cite{team2024gemini,team2025gemini} --- demonstrating that continuing to grow data scale and model capacity yields impressive returns on generalization and robot capabilities. Several initiatives have gathered extensive real-world demonstrations spanning diverse tasks and embodiments, enabling learning from heterogeneous trajectories~\cite{10611477openxembodiment, khazatsky2024droid}. This large-scale heterogeneous real-world data has also been used by recent works to explore cross-embodiment generalization~\cite{zheng2025x,zha2026lap}.

However, these real-world robot datasets are expensive to collect, so the field has increasingly emphasized alternate, cheaper data sources. These alternatives include handheld gripper devices~\cite{chi2024umi,etukuru2024robot}, which still only scale linearly with human effort, and human videos~\cite{singh2024hoi,xu2025flow,bharadhwaj2023generalizable}, which suffer from a domain gap between the human demonstrator and the robot. In contrast, simulation can scale easily and offers extreme controllability in comparison to the real world. Large-scale simulated data has powered generalist robot navigation policies~\cite{internvlan1,eftekhar2025onering,ehsani2024spoc}, but manipulation policies trained on simulated data have either required task-specific post-training, as in \mbox{InternVLA-M1} or \mbox{InternData-A1}~\cite{internvlam1,tian2025interndata}, or remain limited to specific atomic tasks~\cite{deng2025graspvla,wang2025articubot}.

% Owing to the high cost and logistical challenges of real-world data collection, recent research has increasingly emphasized simulation or synthetic datasets. GraspVLA~\cite{deng2025graspvla} explores VLA policies trained on simulated grasping demonstrations, while the InternVLA family (InternVLA-M, InternVLA-A, InternVLA-H/N)~\cite{tian2025interndata} demonstrates large-scale pre-training for manipulation, action planning, navigation, and humanoid control using synthetic trajectories. Similarly, ArticuBot~\cite{wang2025articubot} leverages simulation at scale to generate and train on large demonstration datasets for articulated object manipulation. Additionally, work such as PartInstruct~\cite{Yin2025PartInstructPI} and Infinigen-Articulated~\cite{joshi2025proceduralgenerationarticulatedsimulationready} illustrates the effectiveness of procedurally generated simulation datasets in supporting robot learning research. 

Our work extends this line of research by introducing \trainingdataengine{}, a fully open-source pipeline that enables scalable data generation in simulation across different robots, tasks, and diverse environments, and \trainingdata{}, a large-scale generated dataset of expert manipulation trajectories. By combining procedural scene generation with diverse rigid and articulated assets, our dataset enables training generalist policies that transfer to real-world deployment without any real-world demonstrations or task-specific post-training.

\begin{figure*}[t]
  \centering
  \includegraphics[width=0.9\textwidth]{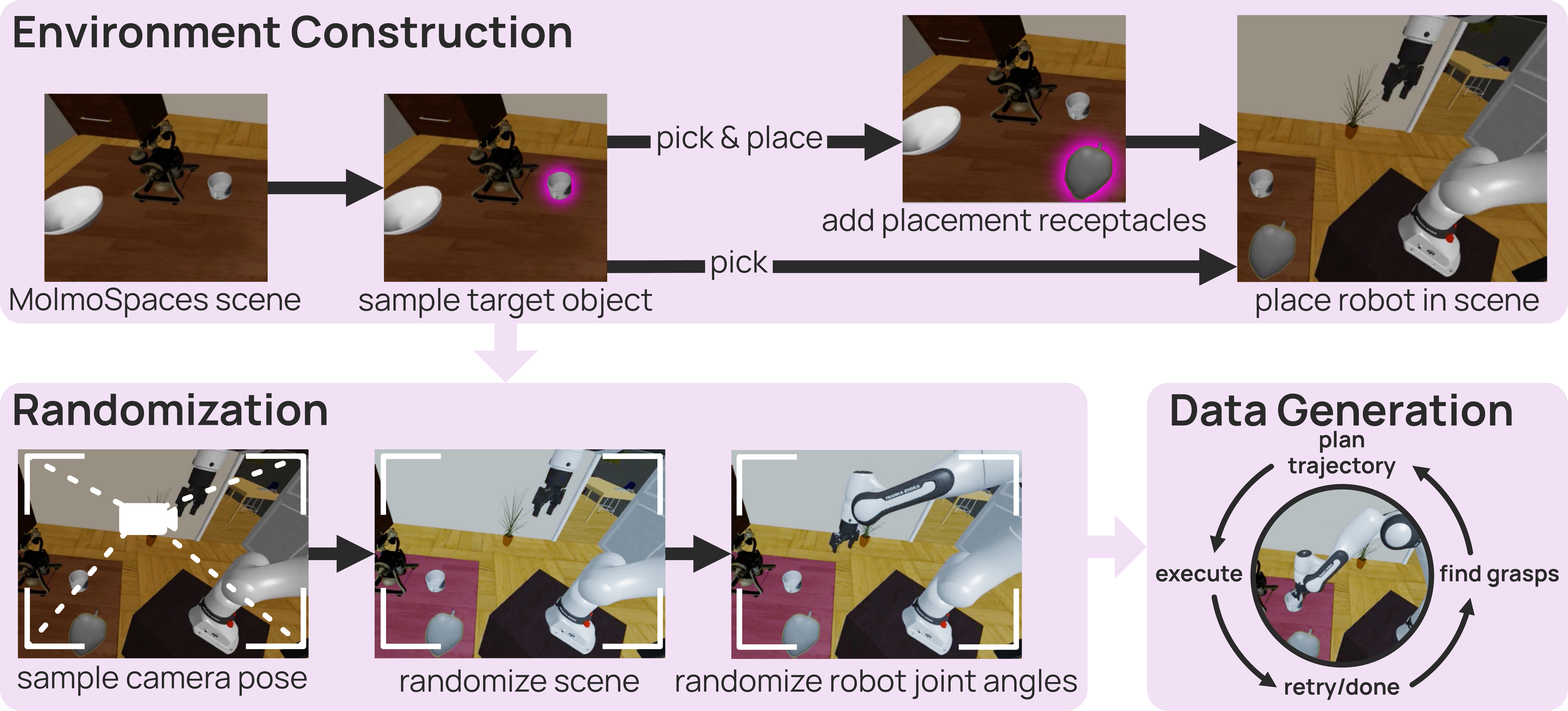}
    \caption{\textbf{\trainingdataengine{}}. Starting from a pre-built MolmoSpaces~\cite{molmospaces2026} scene, we sample task-relevant objects, randomize visual and physical parameters, and iteratively re-plan until a successful trajectory is found.}
    \vspace{-0.3cm}
\label{fig:pipeline}
\end{figure*}

\section{\trainingdataengine{}: A Procedural Data Generation Pipeline}
\label{sec:data}
\label{sec:data_engine}

We introduce \textbf{\trainingdataengine{}}, a scalable data generation pipeline for training robotic manipulation policies, illustrated in Fig.~\ref{fig:pipeline}. Leveraging large-scale procedural generation, massive asset libraries, and domain randomization, we generate demonstration data at scale to make \mbox{\textbf{\trainingdata{}}}, at a fraction of the cost of real-world collection. We present detailed statistics for this novel dataset in Tab.~\ref{tab:dataset_stats} and compare to prior large-scale datasets in Tab.~\ref{tab:dataset_comparison} in the appendix.

% We note that \trainingdataengine{} is inherently constrained by the capabilities of the simulation platform. We focus on rigid body and articulated object manipulation (pick/pick-and-place and door/drawer/cabinet opening), as these are both tractable to model for modern simulators, as well as interesting and challenging tasks still unsolved by modern generalist policies. We hope this contribution can help towards extending simulation data generation to new classes of manipulation such as exceedingly contact-rich or soft-body manipulation.

\subsection{Environments and Assets}
We leverage open-source objects and scenes from MolmoSpaces~\cite{molmospaces2026}, a large collection of procedurally generated indoor environments with realistic and diverse room layouts, object placements, and rigid objects that can be programmatically added to any scene.

\myparagraph{Environment setup.}
Each episode takes place in one of the more than 200k available pre-built MolmoSpaces scenes. The layout, furniture, and static objects remain fixed, and additional manipulable objects are inserted as required for task sampling.

\myparagraph{Asset sourcing.}
Rigid objects for pick-and-place tasks are sourced from iTHOR \cite{Kolve2017AI2THORAI} and Objaverse \cite{Deitke2022ObjaverseAU}, and filtered for manipulability, which includes size and watertight meshes. Objects can be looked up via semantic relevance using the object metadata provided by MolmoSpaces.
% (placement receptacles with bounding boxes of side under 50 cm along the $x$ and $y$ axes and vertical size up to 15 cm, pickup objects with $xy$-plane diagonal less than that for the receptacle) and watertight collider meshes. For task roles like the receptacle target in a pick-and-place task, we additionally ensure semantic relevance by filtering based on the object metadata provided by MolmoSpaces.

\myparagraph{Domain randomization.} We extensively perform domain randomization across three axes: environment randomization, action randomization, and camera perturbation. Furthermore, during model training we also perform image augmentation.

For environment randomization, we randomize lighting (number of lights, positions, type, intensities, colors, and shadow parameters), textures (surface materials are randomized where supported with procedural textures and those sourced from AI2THOR assets~\cite{kolve2017ai2thor}), and dynamics (frictional and inertial properties, joint damping). We also randomize object configurations by sampling initial poses subject to collision and robot reachability constraints.

% Focusing on \textbf{environment randomization}, after object placement, we randomize all visual and physical parameters supported by MuJoCo:
% \begin{itemize}
% \item \textbf{Lighting:} Number of lights ([1--N]), positions, intensities, colors, and shadow properties. We sample both point and directional lights to simulate diverse indoor conditions.
% \item \textbf{Textures:} Surface materials are randomized across placed objects and, where supported, existing scene elements. We sample from procedural textures and real-world texture maps sourced from AI2THOR assets~\cite{kolve2017ai2thor}.
% \item \textbf{Dynamics:} Friction coefficients, object masses, and joint damping are sampled within plausible ranges to encourage robust control policies.
% \end{itemize}

%\rmh{I forget if there's articulated object position randomization} \mayag{no}

% \myparagraph{Pose randomization.}
% Manipulable assets are placed at randomized 6-DoF poses within the environment, subject to collision constraints and reachability from the robot's workspace. We ensure diverse approach angles by sampling asset orientations relative to the robot base.

\subsection{Setup: Robotic Platforms and Tasks}
\label{sec:data_robots}
We generate data for two robot platforms to enable tabletop and mobile manipulation. %Additional robot platforms can be easily added by future work.

\myparagraph{Franka FR3.}
A 7-DoF Franka FR3 arm in the DROID~\cite{khazatsky2024droid} configuration, mounted on a fixed base. Following DROID, data generation and evaluation are run at 15 Hz.

\myparagraph{Rainbow RB-Y1.}
A mobile bimanual manipulator with a holonomic base and actuated 6-DoF torso and 2-DoF head. Each arm is 7-DoF and equipped with a parallel-jaw gripper.
% A mobile manipulator with a holonomic base (3-DoF: $x, y, \theta$), a 6-DoF torso, a 2-DoF head (pan, tilt), and two 7-DoF arms, each equipped with a mechanically coupled parallel-jaw gripper. The base is controlled in planar joint-position mode; the head is passively set at initialization and not actuated during episodes.

% \myparagraph{Robot-related randomization.}
For each episode, we sample initial robot joint positions around the robot-specific nominal home configuration. Additionally, during the episode, the demonstrator's outputted actions are perturbed via proportional action noise injection to ensure state coverage. Further details on joint randomization and action noise injection are provided in App.~\ref{app:datagen_robot_randomization}. Robot cameras are also randomized in both intrinsics and extrinsics, detailed in App.~\ref{sec:data_sensors}.

% \subsection{Task Definitions}

%\myparagraph{Rigid object manipulation.}
We define four rigid-body and two articulated-object manipulation tasks. We evaluate rigid-body tasks on both platforms: \textit{Pick}, \textit{Pick-and-place-into}, \textit{Pick-and-place-next-to}, and \textit{Pick-and-place-color}. We evaluate the articulated-object tasks on the RB-Y1: \textit{Open} and \textit{Open-door}. Please see  App.~\ref{app:tasks} for task setup details.

\subsection{Expert Task and Motion Planners}
\label{sec:data_planners}

For each task, we generate expert demonstrations at scale via scripted motion planners that iteratively sample grasps, verify feasibility, and execute motion for each task phase. Expert demonstrators for the Franka FR3 use IK-based interpolation, and for RB-Y1 use the CuRobo~\cite{sundaralingam2023curobo} motion generator to coordinate the many degrees of freedom with collision-aware motion planning. Pregenerated grasps from MolmoSpaces-Grasp~\cite{molmospaces2026} are used to cheaply sample stable grasps during data generation. Finally, programmed retry behavior provides recovery data, serving to robustify learned policies. Complete details are provided in App.~\ref{app:expert_demonstrations}.

\section{Training MolmoBot Policies on Simulated Data}
\label{sec:training}

\begin{figure*}[t]
  \centering
  \includegraphics[width=0.9\textwidth,trim={3.5cm 0 0 0},clip]{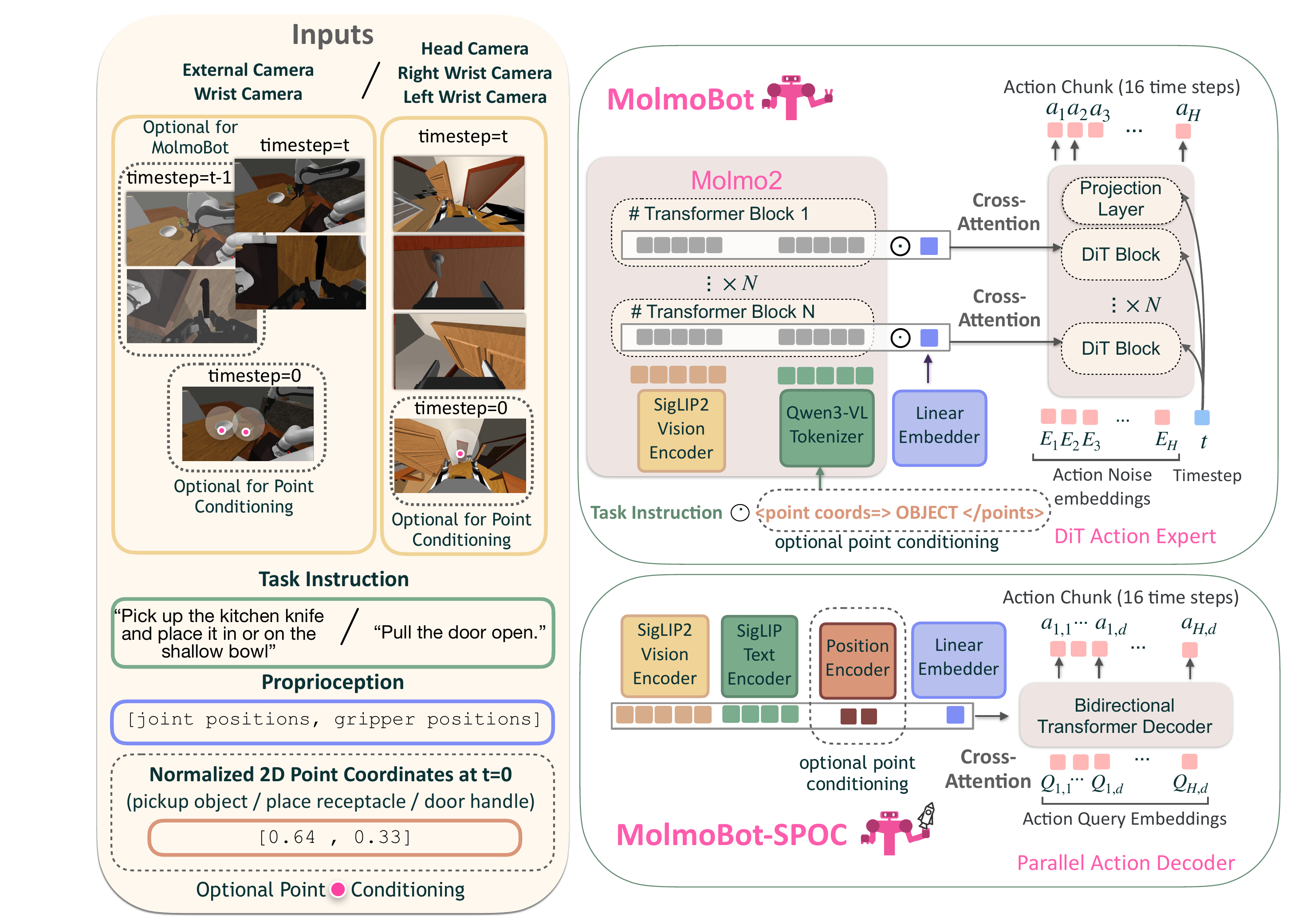}
  \caption{\textbf{Left:} \modelfamily{} policies ingest camera views, proprioception, the task instruction, and optional 2D points, represented as a grounding image and 2D coordinates in that image. \textbf{Top right:} \molmomodel{}, our flagship VLA, uses a Molmo2-4B vision-language backbone and a DiT-based flow matching action head to predict action chunks. \textbf{Bottom right:} \spocmodel{} is a lightweight transformer-based policy, using a transformer decoder to predict an action chunk, conditioned on frozen vision and text embeddings. \paligemmamodel{} (not shown) exactly follows the $\pi_0$~\cite{black2024pi_0} architecture, initialized from PaliGemma~\cite{beyer2024paligemma} weights and a randomly initialized action expert.}
  % \caption{\textbf{Policy architectures.} We train three policy classes on \trainingdata{}. \textbf{Left:} Input observations include RGB images from multiple camera views at the current (and optionally initial) timesteps, proprioceptive state, a language task instruction, and optional 2D point conditioning for specifying target objects or locations. \textbf{Top right:} \molmomodel{} uses a Molmo2 vision-language backbone with a DiTX-based flow matching action head that attends to visual features via cross-attention and predicts action chunks of 16 timesteps. \textbf{Bottom right:} \spocmodel{} uses SigLIP2 vision and text encoders with a bidirectional transformer decoder that processes learned action query embeddings to predict actions in parallel. Both architectures support optional point conditioning. \paligemmamodel{} (not shown) exactly follows the $\pi_0$ architecture~\cite{black2024pi_0} to enable controlled comparison.}
  \label{fig:spoc_flow_arch}
  \vspace{-0.5cm}
\end{figure*}

We train three policy classes on \trainingdata{}, enabling comparison across different architectures and model capacities: \textbf{\molmomodel{}}, \textbf{\paligemmamodel{}}, and \textbf{\spocmodel{}}. %We illustrate \molmomodel{} and \spocmodel{} in Fig.~\ref{fig:spoc_flow_arch}, and defer complete descriptions and details to App.~\ref{app:training}.

\subsection{\molmomodel{}: Our Flagship VLA}
\label{sec:model_molmobot}

\molmomodel{} is our flagship policy, which leverages Molmo2-4B~\cite{clark2026molmo2}, a vision-language model pre-trained on large-scale image-text data. The VLM encodes vision-language observations, which is fed into the flow matching action head along with proprioceptive state, as visualized in Fig.~\ref{fig:spoc_flow_arch}.

\myparagraph{VLM Backbone.} \molmomodel{} uses Molmo2-4B as a backbone. Molmo2-4B is pre-trained with both multi-image and video input, making it well-suited to ingesting not only multiple camera inputs, but also multiple observations over time. Exploiting this functionality, we train \molmomodel{} to ingest up to 2 frames of history, in addition to the current observation. To preserve visual pre-training and avoid overfitting to simulation artifacts, we freeze the SigLIP vision encoder for training.

\myparagraph{Action Head.} The action head is a DiT~\cite{peebles2023scalable} with the same number of layers as the VLM backbone, where each layer cross-attends to the corresponding VLM backbone layer's features. Proprioceptive state is fed to the action head by projecting to the feature space and appending to the VLM feature sequence, while the continuous timestep embedding modulates the features, following \cite{black2024pi_0}. For our experiments, we predict chunks of 16 actions and execute 8 before rerunning inference, following~\cite{zhao2023learning}.

% % \myparagraph{Action representation.} We parameterize actions in joint space using two representations: absolute joint positions and joint position deltas. Both are continuous values representing the target configuration for each joint. At each timestep, the policy predicts targets for all actuated joints, including the gripper.  For the RB-Y1's mobile base, we additionally predict base velocity commands (linear and angular) which are concatenated to the joint action. Joint-space control avoids the computational overhead and potential singularities of inverse kinematics at execution time. We train separate model variants on absolute and delta representations for the Franka FR3 task and compare their performance in Section~\ref{sec:experiments}. We only use delta policies for training the mobile manipulation task.

\myparagraph{Data Sampling.} To induce recovery behavior, the demonstration data includes failures and subsequent retries. To avoid learning suboptimal behavior, we sample frames non-uniformly: retries are up-weighted by $3\times$, successful grasps by $2\times$, and task completion behavior by $2\times$.

\myparagraph{Multi-Timestep Supervision.} The action head has a much lighter compute footprint than the LLM backbone. We leverage this during training by sampling multiple flow matching timesteps to supervise in parallel per data sample, leading to faster convergence. We supervise 8 parallel flow matching timesteps per data sample unless otherwise stated, which we ablate in Sec.~\ref{sec:model_ablations}.

\myparagraph{Training with History.} We train variants of our \molmomodel{} with varying amounts of history timeframes. Our single-frame policy is \molmomodelimg{}, which we train with a batch size of 1024 for 200k steps for static manipulation and 100k for mobile manipulation. Leveraging Molmo2's video pre-training, we also train multi-frame models: \molmomodelmultiframe{1} and \molmomodelmultiframe{2}, which respectively ingest 1 and 2 past observations (in addition to the current). The past observations fed into the model are from $D$ steps ago, with $D=8$ (${\approx}0.5$s) in our experiments. To train multiframe models, we start from \molmomodelimg{} and train the model for 50k additional steps with multiframe observations, keeping all other training parameters unchanged. Being our flagship model, we also refer to \molmomodelmultiframe{1} as \molmomodel{}.

\subsection{\spocmodel: A Lightweight Transformer Policy}

\spocmodel{}, pictured in Fig.~\ref{fig:spoc_flow_arch}, is a transformer-based architecture adapted from \citet{ehsani2024spoc}. With 108M trainable parameters, it is our smallest policy class, and the most amenable to lightweight inference on the edge. Using a transformer decoder, learned action query embeddings cross-attend to encoded observations and task conditioning to produce an action chunk.

\myparagraph{Visual, Language, and Proprioceptive Encoding.} Visual observations are encoded using SigLIP2~\cite{siglip2}, retaining the full sequence of patch tokens. Language task instructions are similarly encoded with the SigLIP text encoder~\cite{siglip}. These token sequences are concatenated with proprioceptive state, which is projected to the token space, and used as input to the policy.

\myparagraph{Point Goal Conditioning.} For tasks that provide spatial goal specification, \spocmodel{} optionally incorporates point-based goal conditioning. 1-2 normalized pixel coordinates are provided depending on the task (2 for PnP, 1 for all else), which are encoded with sinusoidal positional encoding and then projected into the token space. A learned coordinate position embedding is added to each encoded point, and the resulting tokens are concatenated with the other inputs.

\myparagraph{Action Representation.} To enable fast and efficient inference, \spocmodel{} uses a discrete action space. Continuous action values are clipped to the 1st and 99th percentiles, and then discretized using quantile binning with 256 bins, which yields approximately equally populated bins. The action decoder predicts a categorical distribution over the 256 bins independently for each action dimension at every timestep in the chunk, and is supervised with a standard cross-entropy loss.

\myparagraph{Parallel Action Decoding.} Following \citet{zhao2023learning}, \spocmodel{} uses a bidirectional parallel decoder. Given $D\times H$ learned action query embeddings, the decoder predicts $D\times H$ action tokens, where $D$ is the action dimension and $H=16$ is the chunk length. Temporal structure is encoded using sinusoidal positional embeddings applied over a flattened sequence of $D\times H$ positions.

\subsection{\paligemmamodel{}: A Systematic Ablation}

To isolate the effect of \trainingdata{} on real-world VLA performance, we train \paligemmamodel{}, a VLA with  architecture identical to $\pi_0$ \cite{black2024pi_0} but trained entirely on our synthetically generated data. This enables head-to-head comparisons with existing SOTA VLAs, controlling for modeling or architecture changes.

\myparagraph{Architecture.} Following \cite{black2024pi_0}, \paligemmamodel{} uses the Paligemma 3B VLM with a flow-matching action expert. We use the openpi \cite{openpi} codebase for all \paligemmamodel{} modeling code, ensuring equivalence with $\pi_0$.

\myparagraph{Training Protocol.} We train for 200k steps at a batch size of 1024 with a learning rate of $5\cdot10^{-5}$, using a 1k step warmup. All other training parameters (flow matching timestep sampling, other optimizer hyperparameters, etc.) are left as the default values. As with \molmomodel{}, we freeze the vision encoder during training.

\section{Experiments}
\label{sec:experiments}

To illustrate the performance and generality of \modelfamily{} policies and the \trainingdataengine{}, we train and evaluate on multiple tasks in various real-world and simulated settings. Crucially, all models are only trained on sim, with zero task-specific or real-world post-training or fine-tuning. All \modelfamily{} policies have never seen any real-robot data.

\begin{table*}[t]
\newcommand{\ci}[1]{{\color{gray}\scriptsize{$\,\pm$#1}}}
\centering
% \vspace{-0.25cm}
\caption{Evaluation on held-out simulation environments, averaged over 1000 episodes per task, with real-world results from Fig.~\ref{fig:real_results_droid}. For pick-and-place tasks, we report both oracle success (success condition fulfilled at any timestep) and success at end (success condition fulfilled at the final timestep). We additionally report the half-width of the 95\% confidence interval bounds. \textbf{(F)} indicates the use of the Filament renderer, and \textbf{RC} indicates a randomly placed exocentric camera.}
\label{tab:sim_results}
\resizebox{\textwidth}{!}{

\begin{tabular}{@{}l@{\hspace{0.5em}}lc!{\vrule}ccc!{\vrule}ccc!{\vrule}c}
\toprule
&\textbf{Model} & \textbf{Pick v1} & \textbf{Pick v2} & \textbf{Pick v2 (F)} & \textbf{Pick v2 RC (F)} & \textbf{PnP Into (F)} & \textbf{PnP Next-To (F)} & \textbf{PnP Color} & \textbf{Avg.} \\
\midrule
&$\pi_{0.5}$~\cite{black2025pi_05} & 18.1\ci{2.4} & 6.4\ci{1.5} & 7.0\ci{1.6} & 8.0\ci{1.9} & 11.7\ci{2.1}/7.6\ci{1.7} & 8.2\ci{2.2}/6.2\ci{1.9} & 10.4\ci{1.9}/6.7\ci{1.6} & 10.0 \\
&$\pi_{0.5}$-Finetune & 48.0\ci{3.1} & 28.3\ci{2.8} & 25.8\ci{2.7} & 29.7\ci{2.9} & 43.5\ci{6.0}/37.4\ci{5.8} & $\mathbf{28.4}$\ci{3.2}/14.7\ci{2.5} & 48.3\ci{3.1}/38.9\ci{3.0} & 36.0 \\
&StereoVLA~\cite{deng2025stereovla} & 6.6\ci{2.6} & 4.3\ci{1.5} & 1.1\ci{1.0} & N/A & 0 & N/A & 0 & 1.7 \\
&LAP-VLA~\cite{zha2026lap} & 19.4\ci{2.4} & 2.4\ci{1.0} & 3.1\ci{1.1} & 2.7\ci{1.0} & 3.81\ci{1.5}/1.59\ci{1.0} & 6.48\ci{2.8}/3.41\ci{2.1} & 3.1\ci{1.1}/1.5\ci{0.8} & 4.8 \\
&X-VLA~\cite{zheng2025x} & 3.3\ci{1.0} & 0.5\ci{0.5} & 0.7\ci{0.5} & 0.8\ci{0.5} & 0.1\ci{0.2}/0.1\ci{0.2} & 1.9\ci{0.9}/1.0\ci{0.7} & 0.9\ci{0.6}/0.5\ci{0.5} & 1.2 \\

\midrule
\rule{0pt}{2ex}
&\paligemmamodel{} & 66.2\ci{2.9} & 35.7\ci{3.0} & 33.3\ci{2.9} & 39.8\ci{3.1} & 44.7\ci{3.1}/38.2\ci{3.1} & 24.7\ci{3.2}/13.3\ci{2.5} & 46.2\ci{3.1}/40.0\ci{3.1} & 41.5 \\

&\molmomodelimg{} & 92.2\ci{1.7} & 63.5\ci{3.0} & 61.4\ci{3.0} & 62.1\ci{3.0} & 63.0\ci{3.0}/55.0\ci{3.1} & 21.0\ci{2.5}/16.4\ci{2.3} & $\mathbf{67.8}$\ci{2.9}/$\mathbf{60.3}$\ci{3.0} & 61.6 \\

&\molmomodelmultiframe{1}  & $\mathbf{93.5}$\ci{1.5} & $\mathbf{66.8}$\ci{2.9} & $\mathbf{64.0}$\ci{3.0} & $\mathbf{63.7}$\ci{3.0} & $\mathbf{66.4}$\ci{2.9}/$\mathbf{57.7}$\ci{3.0} & 26.4\ci{2.7}/20.2\ci{2.5} & $\mathbf{67.8}$\ci{2.9}/60.0\ci{3.0} & $\mathbf{64.1}$ \\

&\molmomodelmultiframe{2} & 91.3\ci{1.8} & 63.8\ci{3.0} & 59.0\ci{3.0} & 62.7\ci{3.0} & 65.4\ci{2.9}/55.6\ci{3.1} & 28.3\ci{2.8}/$\mathbf{22.6}$\ci{2.6} & 66.1\ci{2.9}/57.3\ci{3.1} & 62.4 \\
\bottomrule
\end{tabular}
}
\end{table*}

\subsection{Zero-Shot Real-World Transfer}
\label{sec:real_eval}

% \subsubsection{Baselines}
For our DROID evaluations, we compare against \pizero{}~\cite{black2024pi_0} and \pizerofive{}~\cite{black2025pi_05}, which are SOTA open-weights manipulation policies for the DROID platform. Both are trained with >10k hours of real-world data and fine-tuned on the DROID dataset, and \pizerofive{} further adds innovations to greatly boost generality. The RB-Y1 enjoys less community adoption than DROID, and  to our knowledge there are no available generalist policies to compare against. We therefore reserve quantitative comparisons for our DROID evaluations.

% \myparagraph{DROID}
% For our DROID evaluations, we compare against $\pi_0\text{-DROID}$ \cite{black2024pi_0} and $\pi_{0.5}\text{-DROID}$ \cite{black2025pi_05}, which are SOTA open-weights manipulation policies for DROID. Both are trained with >10k hours of real-world manipulation demonstrations, and $\pi_{0.5}$ further improves by adding new innovations like subgoal prediction, heterogenous co-training, and more. Therefore, $\pi_0$ is the most closely comparable baseline in terms of modeling decisions, while $\pi_{0.5}$ represents one of the best generalist manipulation policies available today. Note that unlike \trainingdata{}, the data used to pre-train the $\pi$-family models is not publicly available, limiting future reproduction or study by the field at large.

% \myparagraph{RB-Y1}
% The RB-Y1 has significantly less community adoption than DROID. \modelfamily{} policies are therefore, to our knowledge, the first generalizable pick-and-place and articulated object manipulation policies available for the RB-Y1. Lacking other generalist baselines, we reserve quantitative comparisons for our DROID evaluations.

\subsubsection{Static Manipulation Evaluation}

\myparagraph{Setup.} We evaluate all Franka policies on pick-and-place tasks in four real-world environments across two institutions, totaling 120 trials per policy. Evaluation environments are pictured in Fig.~\ref{fig:droid_real_environments}, and further details are provided in App.~\ref{app:real_eval_details}.

% \myparagraph{Setup} We evaluate \modelfamily{} policies and baselines using three different physical DROID platforms, in four real-world environments across two geographical locations and institutions. In each environment, we evaluate each policy on 10 pick-and-place tasks, with 3 trials each. Cumulatively, each policy is evaluated 120 times. Evaluation environments are pictured in Fig.~\ref{fig:droid_real_environments}, and further details on real-world environment and task design are provided in  Sec.~\ref{app:real_eval_details} in the appendix.

% For pick-and-place tasks, given a task prompt (e.g. ``put the banana in the black bowl''), the policy must move the specified object to be stably in or on the given receptacle. If the policy accomplishes this for a reasonable amount of time within the episode horizon (900 steps), the trial is counted as a success. Failure to do so within the episode horizon, or exhibiting unsafe behavior (high-speed collisions, etc.) before success counts as a task failure.

\begin{wrapfigure}{r}{0.5\textwidth}
    \centering
    \vspace{-0.5cm}
    \includegraphics[width=\linewidth]{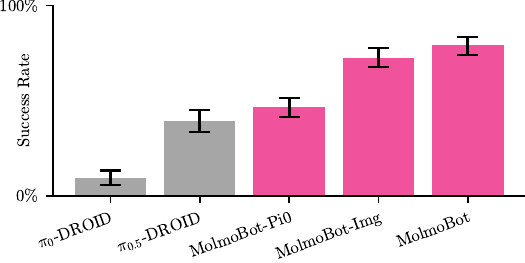}
    \caption{\modelfamily{} policies exhibit strong zero-shot sim2real performance across our 120 real-world DROID evaluations, outperforming SOTA policies trained on large-scale real-world demonstrations. Error bars represent 95\% confidence intervals, estimated via stratified bootstrapping. \molmomodel{} denotes the \molmomodelmultiframe{1} variant.}
    \vspace{-1cm}
    \label{fig:real_results_droid}
\end{wrapfigure}

\myparagraph{Results} In our real-world static manipulation evaluations, \modelfamily{} policies exhibit strong zero-shot sim2real transfer, as illustrated in Fig.~\ref{fig:real_results_droid}. \molmomodel{} and \molmomodelimg{} significantly outperform \pizerofive{} while \paligemmamodel{} is competitive, all without the benefit of \pizerofive{}'s architectural improvements. Additionally, all \modelfamily{} policies perform much better than \pizero{}. Despite having identical architectures, \paligemmamodel{} significantly outperforms \pizero{} in our evaluations, highlighting the value of \trainingdata{}. We further study the dynamics of how this data affects policy behavior in Sec.~\ref{sec:ablations}.

\subsubsection{Mobile Manipulation Evaluation}

\myparagraph{Setup.} We evaluate a \molmomodel{} Door Specialist on a door opening task in three real-world environments, each featuring a distinct pull door with varying textures and handle configurations. Pulling a door open requires precise grasping behavior and coordinated motion to avoid collision.

\myparagraph{Results.} Full evaluation results are presented in Tab.~\ref{tab:door_results}. The robot successfully grasped the door handle in 4 of 9 trials, and in 2 of which successfully opened the door. We note that these evaluations were hampered by hardware faults during rollout. Please see App.~\ref{app:rby1_experiments} for further discussion.

% Some trials were mired by faults triggered by extant hardware issues. In these cases, the e-stop is triggered, preventing policy rollout. We note that upon successfully grasping the handle without triggering a fault, the robot always succeeded in opening the door, which indicates that HW faults are hampering what otherwise would have been successful rollouts. Further failure mode analysis is detailed in App.~\ref{app:rby1_experiments}.

\subsection{Simulation Evaluation}

For further reduced evaluation uncertainty, we turn to systematic simulation evaluation.

\subsubsection{Static Manipulation}
\label{sec:sim_eval}

\myparagraph{Setup.} We evaluate on held-out procedural houses with held-out asset instances, following MolmoSpaces~\cite{molmospaces2026}. We evaluate each task on these houses and report oracle success rate (task completion at any timestep) for all tasks, and additionally success-at-end (task completion at final timestep) for pick-and-place tasks, which captures a policy's tendency to undo task success. To push evaluation settings further OOD from \trainingdata{}, we evaluate some tasks with the Filament renderer, indicated with \textbf{(F)}. Full simulation evaluation details are in App.~\ref{app:sim_eval}.

\myparagraph{Tasks.} We evaluate increasingly difficult tasks on 1000 episode benchmarks, detailed in Tab.~\ref{tab:sim_results}. We evaluate picking tasks with a time budget of 20s under various levels of complexity, which push the evaluations further OOD. This includes Pick v1, which is limited to a select few MSProc object classes, and Pick v2, which uses a diverse set of objects sampled from Objaverse. We additionally evaluate with randomly placed exocentric cameras. For pick-and-place tasks, we evaluate with a time budget of 40s. Further details on task details and success criteria are described in App.~\ref{app:sim_eval}.

\myparagraph{Baselines.}
We compare against existing VLAs: $\pi_{0.5}$~\cite{black2025pi_05} is evaluated both zero-shot and after fine-tuning on \trainingdata{} for 15k steps to adapt it to simulation. We also evaluate StereoVLA~\cite{deng2025stereovla}, LAP-VLA~\cite{zha2026lap}, and X-VLA~\cite{zheng2025x} zero-shot. We note that StereoVLA requires a specific exocentric camera placement and is only trained to place into receptacles, and therefore cannot be evaluated on \textit{Pick v2 RC} or \textit{PnP Next-To}. This is indicated with ``N/A'' and is treated as a 0.

\myparagraph{Results.}
\modelfamily{} policies outperform all baselines across all tasks. \pizerofive{} and $\pi_{0.5}\text{-Finetune}$ are the strongest baselines, averaging 10\% and 36\% success respectively, with the latter ameliorated by fine-tuning on \trainingdata{}. Other VLA baselines (StereoVLA, LAP-VLA, and X-VLA) generally perform poorly in our evaluations, with average success rates of less than 5\%. \modelfamily{} policies, on the other hand, do far better: \paligemmamodel{} averages 41.5\% success while Molmo2-based policies achieve 61\%--64\%. \molmomodelmultiframe{1} performs the best, achieving an average success rate of 64.1\%. Further discussion and additional results are presented in App.~\ref{app:sim_eval}.

\subsubsection{Mobile Manipulation}

\begin{wraptable}{R}{0.55\textwidth}
\caption{Simulation evaluation for RB-Y1 policies on held-out environments. All models evaluated zero-shot without task-specific fine-tuning.}
\label{tab:sim_rby1_results}
\begin{tabular}{l>{\centering\arraybackslash}p{0.8cm}>{\centering\arraybackslash}p{0.8cm}>{\centering\arraybackslash}p{0.8cm}>{\centering\arraybackslash}p{0.8cm}}
\toprule
\textbf{Model} & \textbf{Pick} & \textbf{PnP Into} & \textbf{Open} & \textbf{Open (Door)} \\
\midrule
\molmoshort{} Multitask   & 44.8 & 22.5 & 25.2 & 70.2 \\
\molmoshort{}{} Door & -- & -- & -- & 77.7 \\
\spocshort{} Rigid       & 10.5 & 1.8 & -- & -- \\
\spocshort{} Artic. & -- & -- & 21.8 & 58.8 \\
\bottomrule
\end{tabular}
\end{wraptable}

\myparagraph{Setup.} We evaluate different variants of \modelfamily{} policies on rigid and articulated object manipulation tasks: \textit{Pick}, \textit{PnP Into}, \textit{Open}, and \textit{Open-door}. Specifically, we train and evaluate \molmoshort{} Multitask, trained on all tasks; \molmoshort{} Door, a specialist trained only on door opening; \spocshort{} Rigid, trained only \textit{Pick} and \textit{PnP Into}; and \spocshort{} Artic., trained only on \textit{Open} and \textit{Open-door}. We note that these tasks do not share the same specific environments or starting configurations as those in Sec.~\ref{sec:sim_eval}.

\myparagraph{Results.} As illustrated in Tab.~\ref{tab:sim_rby1_results}, \modelfamily{} policies perform well on a variety of tasks. Evidently, \trainingdataengine{} can generate data for complex high-DOF mobile robots that produce powerful policies. Please refer to App.~\ref{app:rby1_sim_eval} for further discussion.

\subsection{Data Ablations}
\label{sec:ablations}

\FloatBarrier

\begin{figure}[t]
    \centering
    \includegraphics[height=5cm]{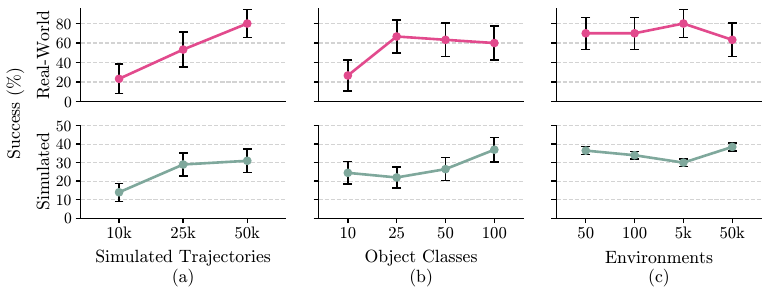}
    \caption{\small Effect of scaling various dimensions of synthetic training data, evaluated in the real world (top) and in simulation (bottom). \textbf{(a)}~Pick performance improves predictably with scale, particularly in real-world evaluations. \textbf{(b)}~Object-class diversity improves simulation performance but not real-world in our evaluations. \textbf{(c)}~Environment diversity does not predictably improve performance. Error bars: 95\% Clopper-Pearson intervals.}
    \label{fig:scaling_combined}
    \vspace{-0.5cm}
\end{figure}

We now proceed to study the various axes of scale and diversity provided by \trainingdata{}, and how they influence downstream performance.

% \subsubsection{Setup}
\myparagraph{Setup.}
All results we report in this section are with the \molmomodelimg{} model architecture, trained only for the \textit{pick} task with various datasets. Simulation evaluations are on \textit{Pick v2}, while real-world evaluations are on the \textit{pick} task, performed with the DROID platform in the workroom environment (Fig.~\ref{fig:droid_real_environments}) on 6 objects with 5 trials per object. We note that each data point in our ablations is derived from a purpose-built dataset of up to 50k trajectories, which is exactly the type of thorough analysis at scale which \trainingdataengine{} aims to enable. Please refer to App.~\ref{sec:ablations_details} for full experimental details.

% \subsubsection{Results}

\myparagraph{Scaling Number of Demonstrations.}
To study how performance changes with the data scale, we vary the total number of demonstrations while keeping the number of environments and object classes fixed. Concretely, we train on 10k, 25k, and 50k demonstrations sampled from the same set of 5k environments and 12.4k object categories. We observe predictable scaling trends as \textit{pick} performance improves with the number of demonstrations (Figure \ref{fig:scaling_combined} (a)).

\myparagraph{Scaling Object Class Diversity.}
\label{sec:ablation_obj_diversity}
Here, we train with a fixed number of 50k trajectories while sampling from 10 to 100 object classes (Fig.~\ref{fig:scaling_combined} (b)). We find that performance improves as expected in simulation, but the performance in real quickly saturates with respect to object class diversity. We hypothesize that the geometric diversity provided by the many object instances within a small number of classes is sufficient for picking, which does not require much semantic reasoning. We note that the policy still performs relatively equally on seen and unseen object classes, implying that the VLM backbone might be responsible for semantic generalization without necessarily requiring robot data.
% because the number of objects in the evaluation is limited and semantically common such as apple and cup.

\myparagraph{Scaling Environment Diversity.}
\label{sec:ablation_env_diversity}
Finally, we vary the number of unique environments in the training set while keeping the total number of demonstrations fixed at 50k. Unexpectedly, we find that increasing the number of unique training environments has little effect on downstream performance (Fig.~\ref{fig:scaling_combined}~(c)). This suggests that geometric and interaction diversity dominate the visual diversity provided by scaling environments. This could be magnified by the strong pretrained visual representations, which might obviate the benefit from added visual diversity. We note that each environment has multiple rooms, which induces substantially more visually distinct workspaces/backgrounds than the number of environments. Extending Fig.~\ref{fig:scaling_combined}~(c) further to the left may be required to more conclusively establish performance trends with respect to environment diversity.
\vspace{-0.2cm}
\section{Conclusion}
\label{sec:conclude}

In this work, we demonstrate that zero-shot transfer of generalist manipulation behaviors from simulation to the real world is not only possible, but state-of-the-art for static manipulation, with demonstrated extension to articulated and mobile manipulation tasks. We introduce \trainingdataengine{} and use it to generate \trainingdata{}, which we use to train a family of generalist policies. Without \emph{any} real-world fine-tuning, our policies demonstrate zero-shot sim-to-real transfer: \molmomodel{} achieves a success rate of 79.2\% in our real-world static manipulation evaluations, more than double that of the strongest baseline, \pizerofive{}, demonstrating broad generalization capability. We release all models, data, and code to enable the community to extend this approach.% Further discussion of limitations and future work are described in Appendix.~\ref{app:limitations}.

\noindent\textbf{Limitations and Future Work:}
\modelfamily{} represents a promising step for scaling up simulation data, but there remain several open challenges. \trainingdataengine{} is constrained to assets and tasks that can be accurately simulated, limiting our focus to rigid body and articulated object manipulation. Extending to contact-rich manipulation or deformable objects poses new difficulties, which may be ameliorated by future advances in physics-based or generative simulators. Additionally, no suitable baselines are currently available for our RB-Y1 evaluations, making them difficult to contextualize within the broader literature. Furthermore, while \modelfamily{} demonstrates the viability of simulation-only training, reliable deployment in the real world may require data from the target environment, on-policy failures and interventions, and more. Rather than supplanting every aspect of real-world data collection, we hope that \trainingdataengine{} serves as a source of high-quality and easy-to-simulate interactions, which can be procedurally scaled up. This would massively reduce the reliance on real-world data collection, which can instead focus on behaviors for which scripted demonstrators are intractable, or tasks which are difficult to simulate. We are excited to see how future work can leverage \trainingdataengine{} and \trainingdata{} to build increasingly robust generalists.
%%DS.05.28: This is a good paragraph, slightly modified but can be improved :)

% Looking forward, our findings inform how future large-scale robot datasets should be collected. If procedural simulation can provide a wealth of diverse data that enables sim-to-real transfer on simpler tasks that can be simulated well, expensive real-world data collection can be focused where it matters most: on complex behaviors in difficult-to-simulate environments. This combines the best of both worlds, the breadth of simulation and the depth of the real world, and offers a more scalable path toward generalist manipulation policies than either data source alone.

%===============================================================================

\clearpage
% The acknowledgments are automatically included only in the final and preprint versions of the paper.
% \acknowledgments{If a paper is accepted, the final camera-ready version will (and probably should) include acknowledgments. All acknowledgments go at the end of the paper, including thanks to reviewers who gave useful comments, to colleagues who contributed to the ideas, and to funding agencies and corporate sponsors that provided financial support.}
\acknowledgments{
This work would not be possible without the support of our colleagues at Ai2:

\begin{itemize}
    \item We thank Christopher Clark and Tanmay Gupta for helpful research discussions and sharing of relevant findings across related projects.
    \item We thank David Albright, Crystal Nam, Kristin Cha, Sophie Lebrecht, Taira Anderson, Kyle Wiggers, Kelsey MacMillan, Katie Morigi, and Megan Bartot for project management, support to robot room, and publicity of MolmoBot.
    \item We thank Sam Skjonsberg, Tucker Wilde, Johann Dahm, Fangzhou Hu, and Caroline Wu for their work on the Ai2 cluster.
\end{itemize}
}

%===============================================================================

\bibliography{main}  % .bib

\clearpage

\appendix

\section{Data Generation}
\subsection{Robot Randomization}
\label{app:datagen_robot_randomization}

\myparagraph{Initial joint-configuration randomization.}
At episode initialization, each move group's joint positions are sampled as $q_0 + \delta$, where $q_0$ is a nominal home configuration and $\delta_i \sim \mathcal{U}(-r_i, r_i)$ with per-joint noise magnitudes $r_i$. For both robots, the arm noise magnitudes are \emph{graduated}: proximal joints receive smaller perturbations and distal joints larger ones. Concretely, the Franka arm uses $\mathbf{r}_\text{arm} = [0.025, 0.05, 0.075, 0.1, 0.125, 0.15, 0.175]$\,rad (chosen via a Jacobian-weighted heuristic to bound TCP displacement to $\leq$\,10\,cm), and each RB-Y1 arm uses $\mathbf{r}_\text{arm} = [0.05, 0.05, 0.075, 0.1, 0.125, 0.15, 0.175]$\,rad. The RB-Y1 additionally randomizes head pan and tilt ($\pm 0.2$\,rad $\approx \pm 11.4^\circ$ each) and gripper aperture ($\pm 0.01$\,rad). Torso and base initial joint positions are not perturbed. %\maxa{move to appendix.}

\myparagraph{Action noise injection.}
During data collection, noise is injected into expert actions to prevent policies from overfitting to exact action replay. The noise is \emph{action-proportional}: its standard deviation scales with the magnitude of the commanded displacement, so stationary commands receive no noise and large motions receive proportionally more.

For arm move groups, noise is applied in TCP space and then mapped back to joint space via the Jacobian pseudo-inverse. Specifically, we compute the commanded TCP displacement $\Delta \mathbf{x} = J \Delta \mathbf{q}$ from the Jacobian $J$ and the joint-space command $\Delta \mathbf{q}$. Position noise is sampled from a truncated Gaussian with $\sigma_\text{pos} = \alpha \| \Delta \mathbf{x}_\text{pos} \|$ and clipped to $\pm 2$\,cm, where $\alpha = 0.1$ is a scale factor. Rotation noise uses $\sigma_\text{rot} = 0.1 \cdot \sigma_\text{pos}$, clipped to $\pm 0.1$\,rad (${\approx}5.7^\circ$). The resulting 6-DoF TCP noise vector $\boldsymbol{\epsilon}_\text{tcp}$ is projected to joint space by solving $J \boldsymbol{\epsilon}_q = \boldsymbol{\epsilon}_\text{tcp}$ in the least-squares sense, and the noisy command is clipped to joint limits.

For the RB-Y1 base, planar noise is applied directly to $(x, y, \theta)$ commands using clipped Gaussians with $\sigma = 0.1 \cdot \| \Delta \mathbf{p} \|$, bounded to $\pm 2$\,cm in position and $\pm 0.05$\,rad ($\approx 2.8^\circ$) in heading. Action noise is disabled during simulated evaluation.

\myparagraph{Gripper handling.}
Gripper close and open commands execute over fixed durations of 0.5s and 0.25s, respectively, followed by a settle period (\texttt{move\_settle\_time}${}= 0.1$\,s for the Franka; up to \texttt{max\_grasping\_timesteps}${}= 5$ control steps for the RB-Y1) during which the arm is held stationary. This simulates real-world grasp settling time and ensures the object is stably grasped before subsequent arm motion resumes.

\subsection{Sensor Configuration}
\label{sec:data_sensors}

After placing objects and applying domain randomization, we configure the robot's sensors for the episode. We describe the camera systems for each platform, followed by additional sensor modalities.

\subsubsection{FR3 camera system}
\label{sec:fr3_cameras}

We generate data for five camera viewpoints for the Franka FR3, to provide diverse viewpoints for tabletop manipulation. In this work, we only train with the wrist camera and one randomized exocentric camera; the rest are present in \trainingdata{} to provide for future work.

\myparagraph{Wrist camera.} A gripper-mounted camera analogous to a ZED Mini, with 52° vertical FOV ($\pm$4° noise). Position is perturbed by $\pm$1.5cm lateral, $\pm$0.5cm vertical, and $\pm$2cm in depth; orientation by $\pm$8° in roll and $\pm$4° in pitch and yaw.

\myparagraph{Fixed shoulder camera.} A robot-mounted exocentric camera positioned at a fixed offset from the robot base, with 71° FOV and light randomization ($\pm$5cm position, $\pm$8° orientation). Placement is constrained to maintain visibility of task objects.

\myparagraph{Randomized exocentric cameras.} Three freely-placed cameras sample positions around the workspace center: two ZED2 analogues (64--72° FOV) and one GoPro analogue (137--140° FOV raw, then warped and cropped). For each camera, we sample distance (0.2--0.8m for ZED2, 0.2--0.5m for GoPro), height (0.05--0.6m above workspace), and azimuth (full 360°). Lookat target is the workspace center with $\pm$10cm noise. Placement is rejected and resampled (up to 20 attempts) if task objects and gripper are not visible.

All FR3 cameras render at $624\times352$, chosen to be close to the real-world resolution of $640\times360$ while keeping both dimensions a multiple of 16 for video encoding.

\subsubsection{RB-Y1 camera system}
\label{sec:rby1_cameras}

The RB-Y1 uses three cameras matching the real robot's sensor configuration, all of which are used for training in this work:

\myparagraph{Head camera.} A head-mounted camera analogous to a GoPro in wide mode, rendered at 1024$\times$576 and cropped to 768$\times$576 (4:3 aspect ratio) in post-processing. We use a vertical FOV of 139° with $\pm$3° noise. Position is perturbed by $\pm$1cm in each axis, orientation by $\pm$4° around each axis, and randomized fisheye warping is applied per-frame during training.

\myparagraph{Wrist cameras.} Left and right wrist-mounted cameras analogous to Intel RealSense D405 sensors. These render at 1024$\times$576 (16:9 aspect ratio) with 58° vertical FOV and $\pm$4° FOV noise. Position noise is $\pm$1.5cm lateral, $\pm$0.5cm vertical, and $\pm$1cm in depth; orientation noise is $\pm$8° in roll and $\pm$4° in pitch and yaw. Depth is recorded  for the benefit of future dataset utility but unused during training.

\subsubsection{Proprioception and additional sensors}

Beyond visual observations, we record proprioceptive state and auxiliary information for analysis and potential future use:

\myparagraph{Robot state.} We record the joint positions and velocities, TCP poses for each gripper, and the robot base pose.

\myparagraph{Action labels.} We record actions in multiple representations: commanded joint positions (absolute and relative to current joint positions), end-effector twist relative to current pose, and absolute end-effector pose. This enables training with different action parameterizations from the same trajectories.

\myparagraph{Task state.} We record the object start and goal poses, grasp state indicators, policy phase, and retry counts from the expert policy.

\myparagraph{Camera parameters.} We record the intrinsic and extrinsic parameters for each camera, enabling projection between 2D and 3D coordinates and potential depth-based augmentation. We also record points in the image frame on objects of interest in all cameras.

Depth images are recorded for RGB-D camera analogues but were not used in any of the training runs reported in this work.

\subsection{Tasks}
\label{app:tasks}

Below we detail specific task descriptions and success criteria for each task we generate data for and evaluate on.

\subsubsection{Rigid object manipulation}
\label{app:rigid_tasks}
\begin{itemize}
    \item \textit{Pick:} Grasp a target object and lift it above its starting height. Success requires that the object is no longer supported by any non-robot surface and has been raised by at least 1\,cm.
    \item \textit{Pick-and-place-into:} Put a target object into a specified receptacle. The task succeeds when at least 50\% of the object's weight is supported by the receptacle, and the receptacle has not been displaced by more than 10\,cm or rotated by more than 45$^\circ$.
    \item \textit{Pick-and-place-next-to:} Place a target object adjacent to a reference object on the same surface. Success requires the surface-to-surface distance in the XY plane to lie within $[0,\, 5]$\,cm and the reference object to remain within 15\,cm of its initial position.
    \item \textit{Pick-and-place-color:} Place an object into a receptacle identified by color (e.g., ``place on the red plate''). Two receptacles identical (except for color) are placed in the scene; success criteria match \textit{pick-and-place}.
\end{itemize}

\subsubsection{Articulated object manipulation}
\label{app:articulated_tasks}
\begin{itemize}
    \item \textit{Open:} Open a nearby articulated object (e.g., cabinet, drawer, oven, dishwasher) to at least 67\% of its joint range.
    \item \textit{Open-door:} Open a nearby hinged door to at least 67\% of its hinge joint range. The instruction is conditioned on the robot's starting pose relative to the door, yielding either ``push the door open'' or ``pull the door open.''
\end{itemize}

\subsection{Expert Demonstrations}
\label{app:expert_demonstrations}

\myparagraph{Grasp sampling and filtering.}
Rather than assuming a fixed grasp pose, we load a large set of pre-computed grasp candidates per object from MolmoSpaces' grasp dataset and progressively filter them:
\begin{enumerate}
    \item \textbf{Candidate loading and ranking:} We load pre-computed 6-DoF grasps for each object, transform them into the world frame (including flipped variants), and rank them by a weighted cost that combines TCP proximity, rotation similarity, vertical alignment, and distance to the object center of mass.
    \item \textbf{Collision filtering:} The top-ranked candidates are tested for gripper--scene collision by placing phantom collision bodies at each candidate pose in MuJoCo and running broadphase collision detection in batches of up to 128.
    \item \textbf{IK feasibility:} Non-colliding candidates are checked for kinematic reachability via batch inverse kinematics (batches of up to 256). The highest-ranked feasible grasp is selected.
\end{enumerate}

\myparagraph{Phase-based trajectory generation.}
Each task is decomposed into a fixed sequence of phases, with motion planned independently per phase.

For \textit{pick-and-place} tasks, the phases are:
\textsc{Pregrasp}~$\to$~\textsc{Grasp}~$\to$~\textsc{Lift}~~$\to$~\textsc{Preplace}$\to$~\textsc{Place}~$\to$~\textsc{Postplace}~$\to$~\textsc{Stow}. For the RB-Y1 demonstrator, the \textsc{Preplace} and \textsc{Place} phases are combined, and we omit the additional \textsc{Stow} phase. In other words, the policy will first move to a pregrasp pose offset along the grasp approach axis, move to the grasp pose and close the gripper, lift the object, move to a pose above the receptacle before lowering to the placement pose and opening the gripper, and finally moving back to the home position.

\textit{Pick-and-place-next-to} and \textit{pick-and-place-color} are the same as \textit{pick-and-place}, differing only in placement pose.

\textit{Pick} tasks proceed similarly to pick-and-place, but terminate after the \textsc{Lift} phase.

% The planner generates a collision-free trajectory to a pre-grasp pose offset along the grasp approach axis, executes a linear approach and gripper close, reverses the approach trajectory to lift, then plans a second collision-free trajectory to the placement pose above the target receptacle.

For the \textit{open} and \textit{open-door} tasks, the phases are:
\textsc{Pregrasp}~$\to$~\textsc{Grasp}~$\to$~\textsc{Articulate}~$\to$~\textsc{Postarticulate}.
After grasping the handle, articulation-specific end-effector waypoints are computed: a circular arc about the hinge axis for revolute joints (doors), or a linear path along the slide axis for prismatic joints (drawers). The planner solves for each waypoint sequentially using IK or trajectory optimization.

\myparagraph{Retry behavior}
Each demonstrator is equipped with retry behavior. While executing a task, if the demonstrator detects a mistake (the object fell out of the grasp, the robot failed to acquire the grasp, etc.) it will reset to the first phase of the trajectory and try again. If more than 3 retries are triggered in an episode, it is terminated and discarded. This explicit retry behavior imbues policies with the ability to handle and recover from mistakes or disturbances.

\myparagraph{Motion planning for RB-Y1 with CuRobo.}
For the RB-Y1, we use CuRobo~\cite{sundaralingam2023curobo} for GPU-accelerated collision-aware trajectory optimization. We model the whole-body kinematics as a 23-DOF chain: a 3-DOF holonomic base (two virtual prismatic joints for planar translation and one continuous joint for yaw), a 6-DOF torso, and two 7-DOF arms. Given a target end-effector pose, CuRobo first solves inverse kinematics using 64 seeds, then computes a collision-free trajectory using 4 trajectory optimization seeds, and finally smooths and interpolates the result to match the simulation control frequency. The robot geometry is approximated by collision spheres, while scene obstacles use mesh-based collision checking with a 0.2m activation distance and self-collision optimization enabled. For phases requiring collision-free motion (e.g., pre-grasp approach, placement), multiple candidate goal poses are evaluated in parallel (default batch size of 4, up to 4 batches), and the trajectory with the least total joint displacement is selected. When a waypoint cannot be reached within a fixed number of control steps, the planner re-plans from the current configuration, up to a maximum of 5 re-planning attempts per phase.

\subsection{Referral Expressions}
\label{app:referral}

In order to sample referral expressions, we first consider task contexts. For example, in a task where the robot is in front of a workbench with some objects on top of it, the context should be the set of objects lying on the workbench surface. In that context, we seek to maximize the contrast between (1) the CLIP similarity between the normalized textual embedding of the referral expression and the normalized visual embedding of the target object, and (2) the CLIP similarity between the referral expression and any of the other objects on the workbench surface. In other words, we do not take into account whether the referral expression would be a better fit for any other object in the scene, as it is the robot's task to infer the correct context given the task setup and the instruction.

The set of possible referral expressions is composed of LLM-generated short descriptions (between 1 and 5 words), synset lemmas, and normalized object category names. We filter valid expressions that produce a CLIP-similarity contrast $\geq 0.03$ while providing a CLIP similarity $\geq 0.1$ and sample via a softmax distribution with temperature $2\cdot10^{-2}$ over the CLIP similarity contrasts for each filtered expression.

\subsection{Train-Time Task Prompt Randomization}
\label{app:prompt_randomization}

To further boost the diversity of the language instructions, we procedurally randomize the task prompt during training.

\myparagraph{Template Randomization}
There are many ways to give the same instruction to the policy. Therefore, at train-time, when sampling data, we sample one of several task prompt templates with varying wording and phrasing for use. For brevity, we defer the full list of task prompt templates to the released code.

\myparagraph{Referral Expression Randomization}
\trainingdata{} saves multiple valid referral expressions for each task-relevant object for each episode. During training, we then sample a referral expression for each object, biasing towards shorter expressions, which we then insert into the sampled prompt template.

\subsection{Dataset Statistics}

Table~\ref{tab:dataset_stats} summarizes \trainingdata{}. We generate 1.7M episodes comprising 295M frames across more than 11k unique target object assets, more than 9k receptacle object assets, and more than 94k environments, which are further modified by the procedural addition of task-relevant per-episode objects. In total, \trainingdata{} is about 12TB of robot experience.

\begin{table*}[t]
\centering
\small
\caption{\trainingdata{} statistics by task. All episodes include RGB observations, proprioceptive state, action labels, and privileged information such as object visibility that the use of simulation affords. The number of assets reflects pickup objects and receptacles for pick-and-place and variants.
}
\label{tab:dataset_stats}
\begin{tabular}{lcccccc}
\toprule
\textbf{Task} & \textbf{Robot} & \textbf{Episodes} & \textbf{Assets (Obj. + Rec.)} & \textbf{Envs.} & \textbf{Avg. Length} & \textbf{Total Length} \\
\midrule
Door-open & RBY1 & 79.0k & -- & 15.1k & 19.6 s & 429 h \\
Open & RBY1 & 46.6k & 217 & 10.5k & 14.8 s & 192 h \\
Pick & RBY1 & 62.3k & 4.2k & 28.0k & 10.7 s & 184 h \\
Pick & Franka & 781.8k & 10.7k & 73.2k & 4.8 s & 1,042 h \\
Pick-and-Place & RBY1 & 14.8k & 2.4k + 183 & 9.3k & 14.0 s & 58 h \\
Pick-and-Place & Franka & 554.2k & 7.0k + 494 & 61.6k & 17.1 s & 2,638 h \\
PnP Next-To & Franka & 182.7k & 931 + 9.0k & 44.9k & 20.1 s & 1,022 h \\
PnP Color & Franka & 28.6k & 3.1k + 183 & 5.3k & 17.4 s & 138 h \\
\midrule
\textbf{Total} & & \textbf{1.7M} & \textbf{11.4k + 9.4k} & \textbf{94.2k} & \textbf{11.7 s} & \textbf{5,704 h} \\
\bottomrule
\end{tabular}
\end{table*}

\myparagraph{Comparison to prior datasets.}
Table~\ref{tab:dataset_comparison} compares \trainingdata{} to prior manipulation datasets.

\begin{table*}[t]
\centering
\small
\caption{Comparison to prior manipulation datasets. \trainingdata{} provides substantially more episodes and environment diversity through procedural generation.
}
\label{tab:dataset_comparison}
\begin{tabular}{lccrcccc}
\toprule
\textbf{Dataset} & \textbf{Source} & \textbf{Episodes} & \textbf{Hours} & \textbf{Uniq. Envs.} & \textbf{Tasks} & \textbf{Embod.} & \textbf{Mob. Manip.} \\
\midrule
DROID \cite{khazatsky2024droid}                & Real     & 76k   & 350    & 564    & 86  & 1  & \xmark \\
OXE \cite{o2024open}             & Real     & 1.4M   & $\sim$2k     & 311     & 527 & 22 & \cmark \\
AgiBot-World \cite{agibotworldcolosseo2025}    & Real     & 1M+   & 2,976  & 100+   & 217 & 1  & \cmark \\
RoboMimic \cite{robomimic2021}                 & Sim      & $\sim$1k & --  & 1      & 5   & 1  & \xmark \\
MimicGen \cite{mandlekar2023mimicgen}          & Sim      & 50k+  & --     & --     & 18  & 1  & \xmark \\
InternData-A1 \cite{tian2025interndata}        & Sim      & 630k  & 7,433  & 227    & 70  & 4  & \xmark \\
RoboCasa-365 \cite{Nasiriany2026robocasa365}   & Sim      & 500k  & 2,200  & 2,500  & 365 & 1  & \cmark \\
\midrule
MB-Data (Ours)                         & Sim      & 1.7M  & 5,704  & 94.2k & 8   & 2  & \cmark \\
\bottomrule
\end{tabular}
\end{table*}

\myparagraph{Generation throughput.}
%% Estimate from pick datagen from job/total_episodes over time
A key advantage of simulation-based data generation is scalability. Using 100 NVIDIA A100 80GB GPUs, we generate approximately 260 average / 660 peak successful episodes per GPU-hour (with about 50\% of the time spent on rollouts and the rest on scene setup and task sampling) or, equivalently, more than 88 hours of robot experience per hour of wall-clock time. We reject unsuccessful rollouts, so this is lower than the raw generation throughput, since we rejection sample at the rate of demonstration success. The combined planner + demonstrator success rate typically ranges from ${\approx}25\%$ for Open (RB-Y1) to ${\approx}80\%$ for Pick (Franka). At commodity cloud GPU rates for A100s, this comes out to \$0.01--\$0.02 per trajectory, or \$2--\$5 per hour of robot data.  The full \trainingdata{} dataset was generated in approximately 6,500 GPU-hours. This represents a near 2.6×
% cost reduction
data throughput\footnote{Using ALOHA \cite{zhao2023learning} as reference, where the effective real-time factor of a single human demonstrator is 1/3 for tasks of similar duration to ours, due to episode reset overhead or operator mistakes.} compared to real-world data collection at equivalent scale with human demonstrators, enabling rapid iteration on data composition and task design as well as rapid adoption of new robotics platforms.

\section{Models}

In Sec.~\ref{sec:training} we introduced our three policy classes: \molmomodel{}, \paligemmamodel{}, and \spocmodel{}. We now proceed to describe additional details for policy implementation.

\subsection{Optional point conditioning}

\modelfamily{} policies can optionally take in point conditioning in addition to a language instruction, helping with goal specification and alleviating the ambiguity of language

\myparagraph{\molmomodel{}.} For \molmomodel{}, the point conditioning takes the form of a ``grounding frame'' (i.e. the first frame of the episode) along with one or more normalized 2D pixel coordinates on that image. The grounding frame is passed to the model as an additional image input, and the 2D coordinates are expressed as XML in the format \verb|<point coord=...>{object_name}</point>| and appended to the language task instruction.

\myparagraph{\spocmodel{}.} \spocmodel{} point conditioning is explained in Sec.~\ref{sec:training}.

Note that our DROID models and evaluation do not use this point conditioning; \molmomodel{}-Multitask, \molmomodel{}-Door, and \spocmodel{}-Artic. in Tab.~\ref{tab:sim_rby1_results} are the only policies that use it.

\subsection{Data mixing} 

\begin{table}[htb]
    \centering
    \caption{Multitask data mixture for all \modelfamily{} Franka FR3 policies. The data mix is selected to ensure all coverage of each of the individual task's training set.}    \begin{tabular}{lr}
    \toprule
    \textbf{Task} & \textbf{Sampling Ratio} \\
    \midrule
    Pick & 20\% \\
    Pick-and-place Fixed Height & 10\% \\
    Pick-and-place Random Height & 35\% \\
    Pick-and-place-next-to & 20\% \\
    Pick-and-place-color & 15\% \\
    \bottomrule
    \end{tabular}
    \label{tab:training_datamix}
\end{table}

% \FloatBarrier

\begin{table}[htb]
    \centering
    \caption{Multitask data mixture for all \modelfamily{} and \spocmodel{} RB-Y1 policies.}
    \begin{tabular}{lcccc}
    \toprule
    \textbf{Model} & \textbf{Open} & \textbf{Door-open} & \textbf{Pick} & \textbf{Pick-and-place}\\
    \midrule
    \modelfamily{} Multitask  & 20\%  & 20\%  & 30\%  & 30\% \\
    \modelfamily{} Door Specialist & - & 100\% & - & - \\
    \spocmodel{} Rigid        & -  & -  & 50\% & 50\% \\
    \spocmodel{} Articulated  & 45\% & 55\% & -  & -  \\
    \bottomrule
    \end{tabular}
    \label{tab:training_datamix_rby1}
\end{table}

Table~\ref{tab:training_datamix} details the different training sets we use to train all the Franka FR3 policies. Pick-and-place Random Height comprises of pick and place tasks with the robot position initialized at random heights, while Pick-and-place Fixed Height initializes the model at the default DROID position~\cite{khazatsky2024droid}. Training with randomized heights makes the model more robust to inference time variations. Pick-and-place-color trains the model to attend to the color attribute in input task and Pick-and-place-next-to training helps improve the models spatial understanding. Table \ref{tab:training_datamix_rby1} details the various data mixtures used to train RB-Y1 policies. Door-open is a specialized door opening task, while the Open task includes training samples to open cabinets and drawers.

\subsection{Data augmentation}

We use image augmentation to
%train the model
improve our models' sim to real transfer. Specifically, we use ColorJitter, GaussianBlur, RandomPosterize, RandomSharpness and RandomGrayscale with different probabilities. Furthermore, we add prompt randomization during training to make the robot robust to inference time variation in language instruction. Further details on prompt randomization are detailed in Sec.~\ref{app:prompt_randomization}.

\subsection{DROID Input/Output}

For all \modelfamily{} models trained for the DROID platform, we train on 2 of the cameras detailed in Sec.~\ref{sec:fr3_cameras}: the wrist ZED Mini camera, and one randomized exocentric ZED 2 camera. In addition to images, the current joint angles are also given to the model. All \modelfamily{} DROID models output actions as absolute joint position commands.

\subsection{RB-Y1 Input/Output}

For all \modelfamily{} models trained for the RB-Y1 platform, we train on all 3 generated camera views: the head and both wrist cameras. \spocmodel{} Articulated uses the repeated first frame of the trajectory and a normalized image point to ground the task spatially. The point is sampled from ten candidates that are on the desired task object to be manipulated. \molmomodel{} uses similar point conditioning for the door and articulated tasks. We also provide current joint angles. All RB-Y1 models output torso and arm actions as delta joint position commands, and mobile base commands as linear and angular offsets from the current pose.

\section{Experiments}
\subsection{DROID Evaluation Environments}
\label{app:real_eval_details}

\begin{figure*}[htb]
    \centering
\includegraphics[width=0.24\linewidth,clip,trim=0 0 0 0]{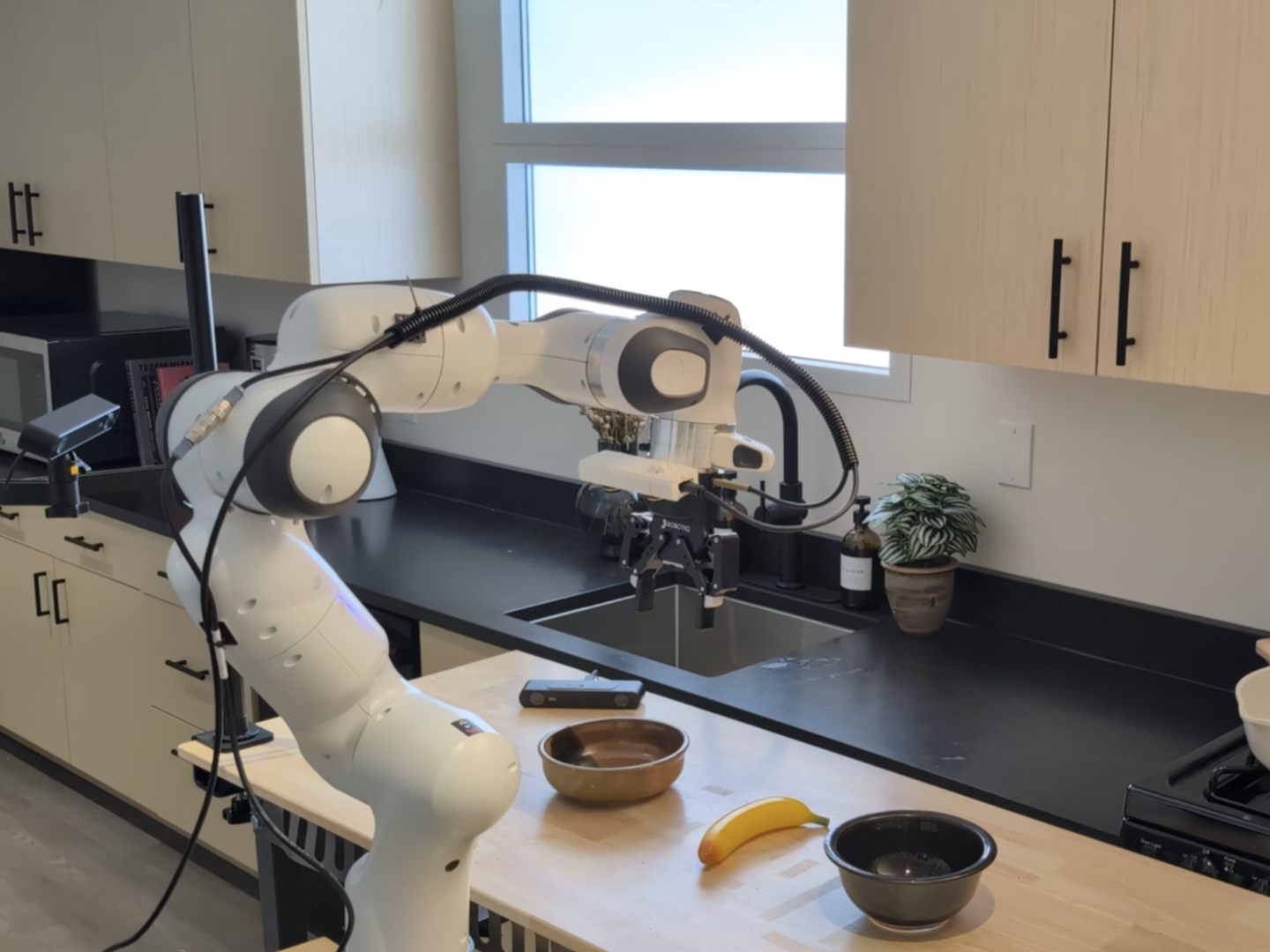}
    \includegraphics[width=0.24\linewidth,clip]{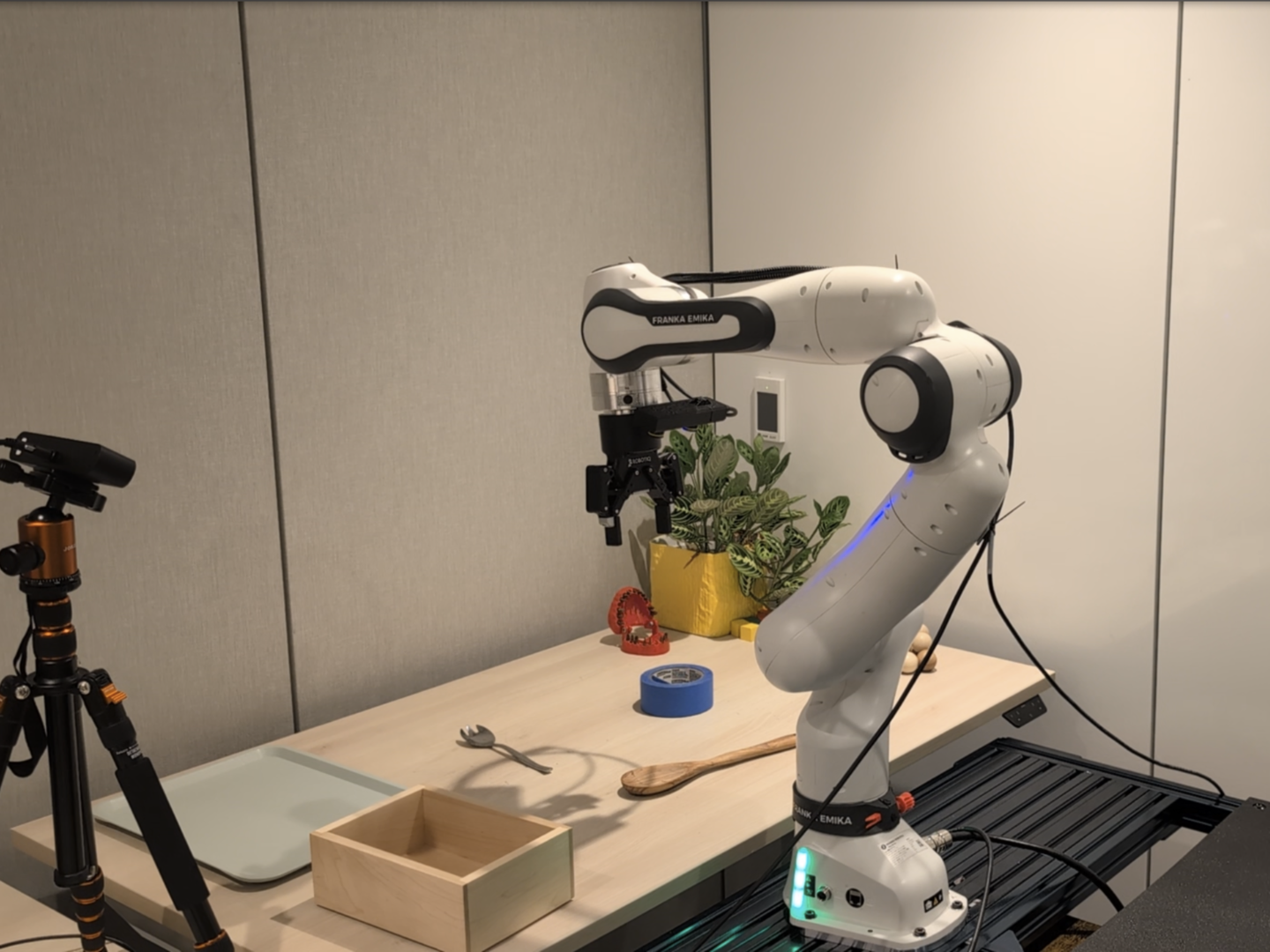}
    \includegraphics[width=0.24\linewidth,clip]{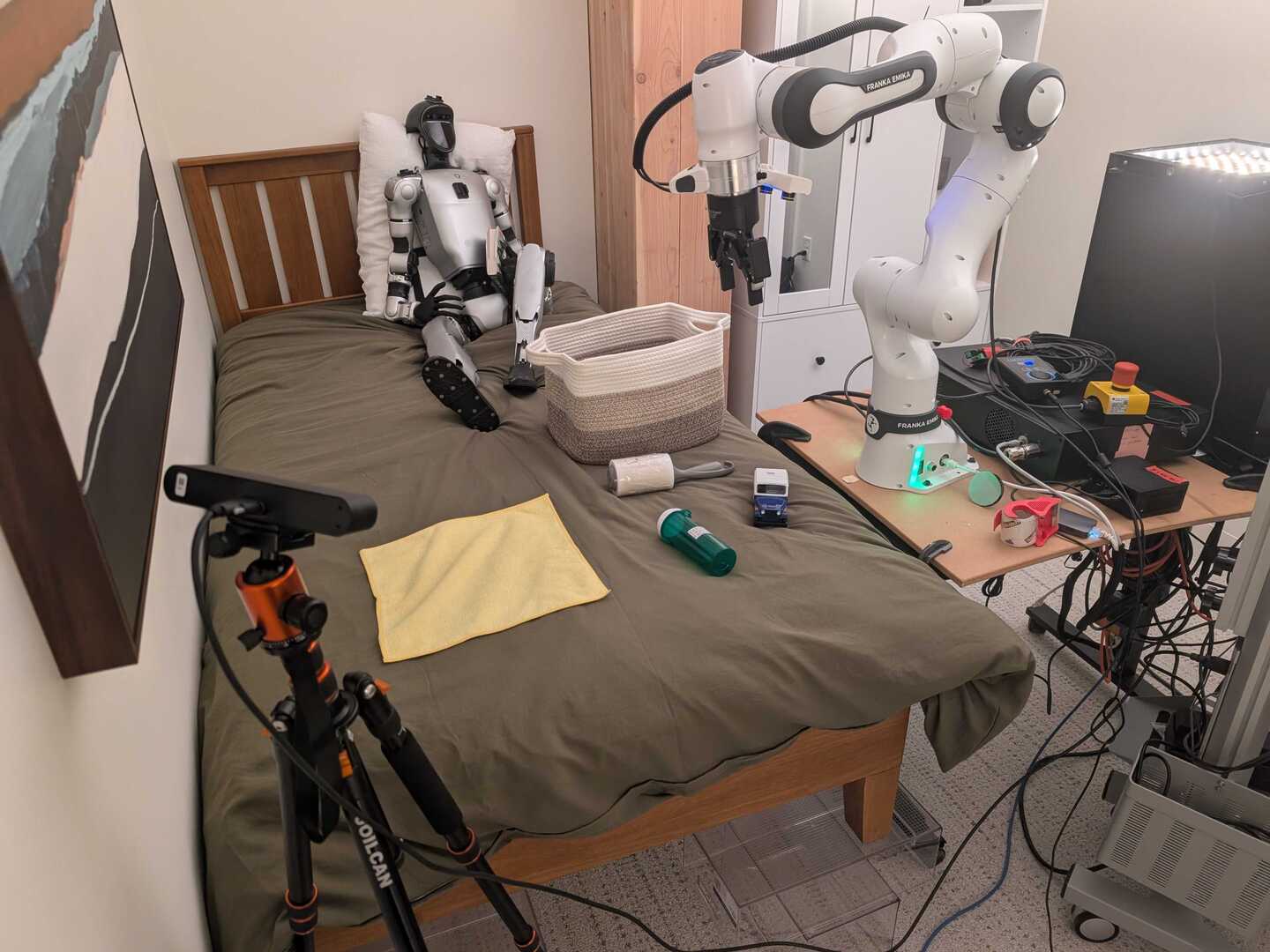}
    \includegraphics[width=0.24\linewidth,clip]{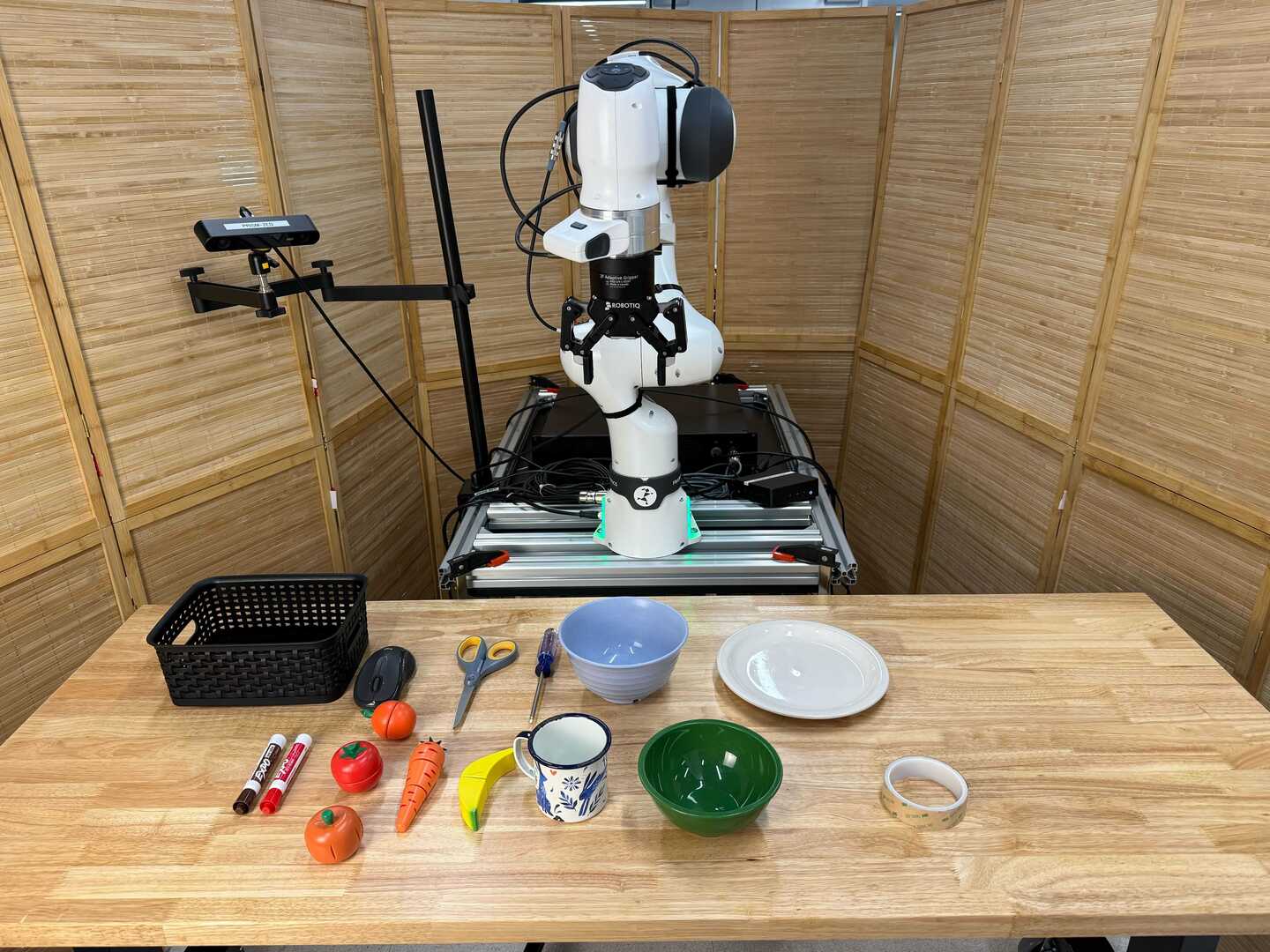}
    \caption{Real-world environments for our DROID evaluations. From left to right: kitchen, workroom, bedroom, office.}
    \label{fig:droid_real_environments}
\end{figure*}

Below we describe each real-world environment used for DROID evaluations in the real world, pictured in Fig.~\ref{fig:droid_real_environments}. Each environment's exocentric agent view is different. Specific task prompts for each task are detailed in Tab.~\ref{tab:droid_task_prompts}. We also present the full results of every real-world evaluation trial for each policy on each task in each environment in Tab.~\ref{tab:droid_pertask_results}.

For pick-and-place tasks, given a task prompt (e.g. ``put the banana in the black bowl''), the policy must move the specified object to be stably in or on the given receptacle. If the policy accomplishes this for a reasonable amount of time within the episode horizon (900 steps), the trial is counted as a success. Failure to do so within the episode horizon, or exhibiting unsafe behavior (high-speed collisions, etc.) before success counts as a task failure.

\myparagraph{Kitchen} The kitchen environment consists of 4 objects (mug, computer mouse, apple, banana) and 2 receptacles (brown bowl, black bowl). The receptacles are placed to the left and right of the robot, while the objects are either placed in between the bowls (``easy'' placement) or further from the workspace center, closer to the wrong bowl (``hard'' placement). For each of these $4\times2=8$ tasks, the target receptacle is the brown bowl. For the final 2 tasks, all objects are placed onto the table, and the policy must put the mug into each of the receptacles.

\myparagraph{Workroom} The workroom environment consists of 5 objects (tape, wooden spoon, timer, copper mug, blue mug) and 2 receptacles (tray, box) on the left of the workspace. Each object must be placed into each receptacle for a total of 10 tasks. When evaluating on the mugs or timer, these objects are placed together on the table in the middle of the workspace. When evaluating on the tape or spoon, they are placed along with an additional spork (a distractor) in the middle of the workspace.

\myparagraph{Bedroom} The bedroom environment consists of 4 objects (pill bottle, lint roller, banana, tennis ball) and 2 receptacles (towel, basket). Each object is placed into each receptacle for $4\times2=8$ tasks, and for the final 2 tasks the policy must put the banana into each receptacle, but with a cluttered workspace. This environment notably does not feature a table as a support surface, but instead a bed, which tests robustness environment diversity.

\myparagraph{Office} The office environment features 8 objects (knife, banana, marker, scissors, carrot, screwdriver, computer mouse, mug) and 7 receptacles (cutting board, plate, mug, green bowl, blue bowl, basket, box), with multiple object configurations with varying amounts of clutter and distractors. Evaluations in this environment were conducted in an entirely different institution and geographical location, illustrating \modelfamily{} policies' ability to get up and running in completely new settings.

\begin{table*}[t]
\centering
\small
\begin{tabular}{l l p{5.5cm}}
\toprule
\textbf{Benchmark} & \textbf{Task ID(s)} & \textbf{Task Prompt} \\
\midrule
\multirow{10}{*}{Workroom} & \texttt{spoon\_tray} & ``put the wooden spoon on the light blue tray'' \\
 & \texttt{spoon\_box} & ``put the wooden spoon in the wooden box'' \\
 & \texttt{tape\_tray} & ``put the blue tape on the light blue tray'' \\
 & \texttt{tape\_box} & ``put the blue tape in the wooden box'' \\
 & \texttt{blue\_mug\_tray} & ``put the blue mug on the light blue tray'' \\
 & \texttt{blue\_mug\_box} & ``put the blue mug in the wooden box'' \\
 & \texttt{copper\_mug\_tray} & ``put the copper mug on the light blue tray'' \\
 & \texttt{copper\_mug\_box} & ``put the copper mug in the wooden box'' \\
 & \texttt{timer\_tray} & ``put the green timer on the light blue tray'' \\
 & \texttt{timer\_box} & ``put the green timer in the wooden box'' \\
\midrule
\multirow{5}{*}{Kitchen} & \texttt{apple\_easy, apple\_hard} & ``put the apple in the brown bowl'' \\
 & \texttt{clutter\_brown, mug\_easy, mug\_hard} & ``put the mug in the brown bowl'' \\
 & \texttt{banana\_easy, banana\_hard} & ``put the banana in the brown bowl'' \\
 & \texttt{mouse\_easy, mouse\_hard} & ``put the computer mouse in the brown bowl'' \\
 & \texttt{clutter\_black} & ``put the mug in the black bowl'' \\
\midrule
\multirow{8}{*}{Bedroom} & \texttt{pills\_towel} & ``put the pill bottle on the yellow towel'' \\
 & \texttt{pills\_basket} & ``put the pill bottle in the basket'' \\
 & \texttt{roller\_towel} & ``put the lint roller on the yellow towel'' \\
 & \texttt{roller\_basket} & ``put the lint roller in the basket'' \\
 & \texttt{banana\_towel, clutter\_towel} & ``put the banana on the yellow towel'' \\
 & \texttt{banana\_basket, clutter\_basket} & ``put the banana in the basket'' \\
 & \texttt{ball\_towel} & ``put the tennis ball on the yellow towel'' \\
 & \texttt{ball\_basket} & ``put the tennis ball in the basket'' \\
\midrule
\multirow{10}{*}{Office} & \texttt{knife\_board} & ``put the knife on the cutting board'' \\
 & \texttt{banana\_plate} & ``move the toy banana on the plate'' \\
 & \texttt{marker\_mug} & ``put the marker into the mug'' \\
 & \texttt{scissors\_bowl} & ``pick up the scissor and place it inside the bowl'' \\
 & \texttt{carrot\_basket} & ``put the carrot into the basket'' \\
 & \texttt{knife\_green\_bowl} & ``pick up the knife and place it inside the green bowl'' \\
 & \texttt{screwdriver\_blue\_bowl} & ``put the screwdriver in the blue bowl'' \\
 & \texttt{mouse\_blue\_bowl} & ``place the mouse into the blue bowl'' \\
 & \texttt{mug\_bowl} & ``grasp the mug and put it inside the bowl'' \\
 & \texttt{marker\_box} & ``pick up the red marker and place it into the box'' \\
\bottomrule
\end{tabular}
\caption{Task prompts for each task in each benchmark for our real-world DROID pick-and-place evaluations.}
\label{tab:droid_task_prompts}
\end{table*}

\begin{table*}[t]
\centering
\small
\setlength{\tabcolsep}{3pt}
\begin{tabular}{l l  c c c c c c c c c c  c}
\toprule
\textbf{Benchmark} & \textbf{Policy}  & \rotatebox{70}{\scriptsize Spoon Tray} & \rotatebox{70}{\scriptsize Spoon Box} & \rotatebox{70}{\scriptsize Tape Tray} & \rotatebox{70}{\scriptsize Tape Box} & \rotatebox{70}{\scriptsize Blue Mug Tray} & \rotatebox{70}{\scriptsize Blue Mug Box} & \rotatebox{70}{\scriptsize Copper Mug Tray} & \rotatebox{70}{\scriptsize Copper Mug Box} & \rotatebox{70}{\scriptsize Timer Tray} & \rotatebox{70}{\scriptsize Timer Box} & \textbf{Avg} \\
\midrule
\multirow{5}{*}{\textbf{Workroom}} & \molmomodel & 3/3 & 2/3 & 3/3 & 3/3 & 3/3 & 3/3 & 3/3 & 1/3 & 3/3 & 3/3 & \textbf{90\%} \\
 & \molmomodelimg & 3/3 & 3/3 & 3/3 & 3/3 & 3/3 & 2/3 & 2/3 & 0/3 & 3/3 & 1/3 & 77\% \\
 & \paligemmamodel & 1/3 & 0/3 & 3/3 & 3/3 & 3/3 & 3/3 & 2/3 & 1/3 & 0/3 & 2/3 & 60\% \\
 & \pizerofive{} & 2/3 & 2/3 & 2/3 & 0/3 & 1/3 & 1/3 & 0/3 & 0/3 & 0/3 & 0/3 & 27\% \\
 & \pizero{} & 0/3 & 0/3 & 1/3 & 0/3 & 0/3 & 0/3 & 0/3 & 0/3 & 0/3 & 0/3 & 3\% \\
\midrule
 &  & \rotatebox{70}{\scriptsize Apple Easy} & \rotatebox{70}{\scriptsize Apple Hard} & \rotatebox{70}{\scriptsize Mug Easy} & \rotatebox{70}{\scriptsize Mug Hard} & \rotatebox{70}{\scriptsize Banana Easy} & \rotatebox{70}{\scriptsize Banana Hard} & \rotatebox{70}{\scriptsize Mouse Easy} & \rotatebox{70}{\scriptsize Mouse Hard} & \rotatebox{70}{\scriptsize Clutter Brown} & \rotatebox{70}{\scriptsize Clutter Black} & \textbf{Avg} \\
\midrule
\multirow{5}{*}{\textbf{Kitchen}} & \molmomodel & 1/3 & 2/3 & 3/3 & 3/3 & 3/3 & 3/3 & 3/3 & 3/3 & 0/3 & 0/3 & 70\% \\
 & \molmomodelimg & 3/3 & 3/3 & 2/3 & 3/3 & 3/3 & 3/3 & 3/3 & 2/3 & 3/3 & 1/3 & \textbf{87\%} \\
 & \paligemmamodel & 2/3 & 3/3 & 2/3 & 0/3 & 3/3 & 1/3 & 3/3 & 2/3 & 0/3 & 0/3 & 53\% \\
 & \pizerofive{} & 3/3 & 0/3 & 2/3 & 1/3 & 3/3 & 2/3 & 3/3 & 1/3 & 2/3 & 2/3 & 63\% \\
 & \pizero{} & 0/3 & 1/3 & 0/3 & 0/3 & 1/3 & 0/3 & 3/3 & 1/3 & 0/3 & 0/3 & 20\% \\
\midrule
 &  & \rotatebox{70}{\scriptsize Pills Towel} & \rotatebox{70}{\scriptsize Pills Basket} & \rotatebox{70}{\scriptsize Roller Towel} & \rotatebox{70}{\scriptsize Roller Basket} & \rotatebox{70}{\scriptsize Banana Towel} & \rotatebox{70}{\scriptsize Banana Basket} & \rotatebox{70}{\scriptsize Ball Towel} & \rotatebox{70}{\scriptsize Ball Basket} & \rotatebox{70}{\scriptsize Clutter Towel} & \rotatebox{70}{\scriptsize Clutter Basket} & \textbf{Avg} \\
\midrule
\multirow{5}{*}{\textbf{Bedroom}} & \molmomodel & 3/3 & 3/3 & 3/3 & 0/3 & 3/3 & 2/3 & 3/3 & 3/3 & 3/3 & 3/3 & \textbf{87\%} \\
 & \molmomodelimg & 0/3 & 0/3 & 1/3 & 2/3 & 3/3 & 3/3 & 3/3 & 2/3 & 3/3 & 3/3 & 67\% \\
 & \paligemmamodel & 2/3 & 0/3 & 2/3 & 0/3 & 0/3 & 0/3 & 1/3 & 0/3 & 0/3 & 2/3 & 23\% \\
 & \pizerofive{} & 0/3 & 0/3 & 0/3 & 0/3 & 1/3 & 0/3 & 0/3 & 0/3 & 2/3 & 0/3 & 10\% \\
 & \pizero{} & 0/3 & 0/3 & 0/3 & 0/3 & 0/3 & 0/3 & 0/3 & 0/3 & 0/3 & 0/3 & 0\% \\
\midrule
 &  & \rotatebox{70}{\scriptsize Knife Board} & \rotatebox{70}{\scriptsize Banana Plate} & \rotatebox{70}{\scriptsize Marker Mug} & \rotatebox{70}{\scriptsize Scissors Bowl} & \rotatebox{70}{\scriptsize Carrot Basket} & \rotatebox{70}{\scriptsize Knife Green Bowl} & \rotatebox{70}{\scriptsize Screwdriver Blue Bowl} & \rotatebox{70}{\scriptsize Mouse Blue Bowl} & \rotatebox{70}{\scriptsize Mug Bowl} & \rotatebox{70}{\scriptsize Marker Box} & \textbf{Avg} \\
\midrule
\multirow{5}{*}{\textbf{Office}} & \molmomodel{} & 2/3 & 3/3 & 1/3 & 2/3 & 2/3 & 2/3 & 3/3 & 2/3 & 3/3 & 1/3 & \textbf{70\%} \\
 & \molmomodelimg{} & 2/3 & 1/3 & 0/3 & 1/3 & 3/3 & 1/3 & 3/3 & 2/3 & 3/3 & 2/3 & 60\% \\
 & \paligemmamodel{} & 1/3 & 3/3 & 0/3 & 0/3 & 0/3 & 1/3 & 3/3 & 2/3 & 3/3 & 2/3 & 50\% \\
 & \pizerofive{} & 2/3 & 3/3 & 1/3 & 1/3 & 1/3 & 1/3 & 2/3 & 1/3 & 3/3 & 2/3 & 57\% \\
 & \pizero{} & 0/3 & 0/3 & 0/3 & 0/3 & 1/3 & 1/3 & 1/3 & 1/3 & 0/3 & 0/3 & 13\% \\
\bottomrule
\end{tabular}
\caption{DROID real-world evaluation results on each task in each environment. Each cell shows successes out of 3 trials.}
\label{tab:droid_pertask_results}
\end{table*}

\subsection{Mobile Manipulation Experiments}
\label{app:rby1_experiments}

\begin{figure*}[t]
    \centering
\includegraphics[width=0.24\linewidth,height=2.5cm,clip,trim=0 0 0 0]{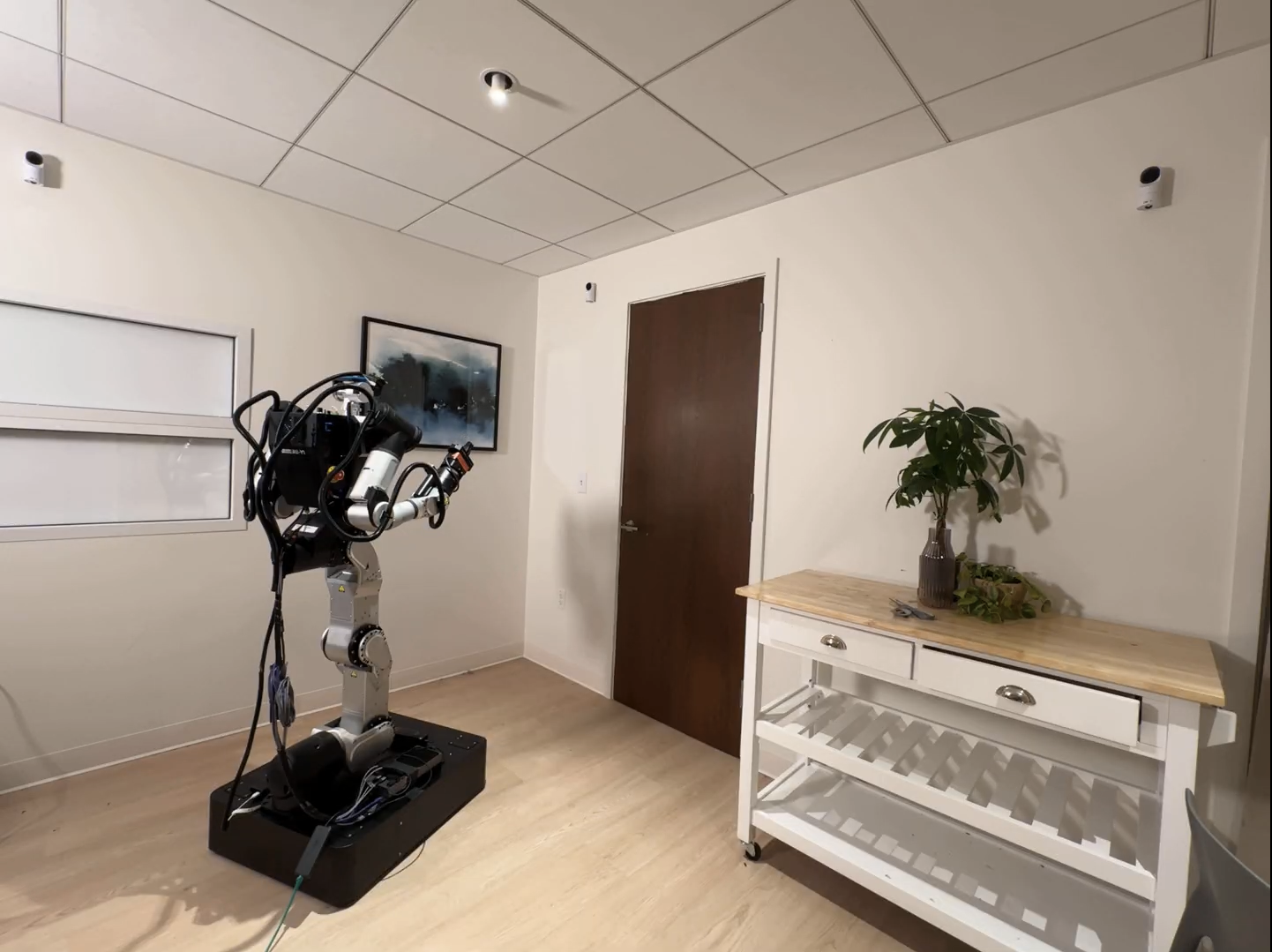}
    \includegraphics[width=0.24\linewidth,height=2.5cm,clip]{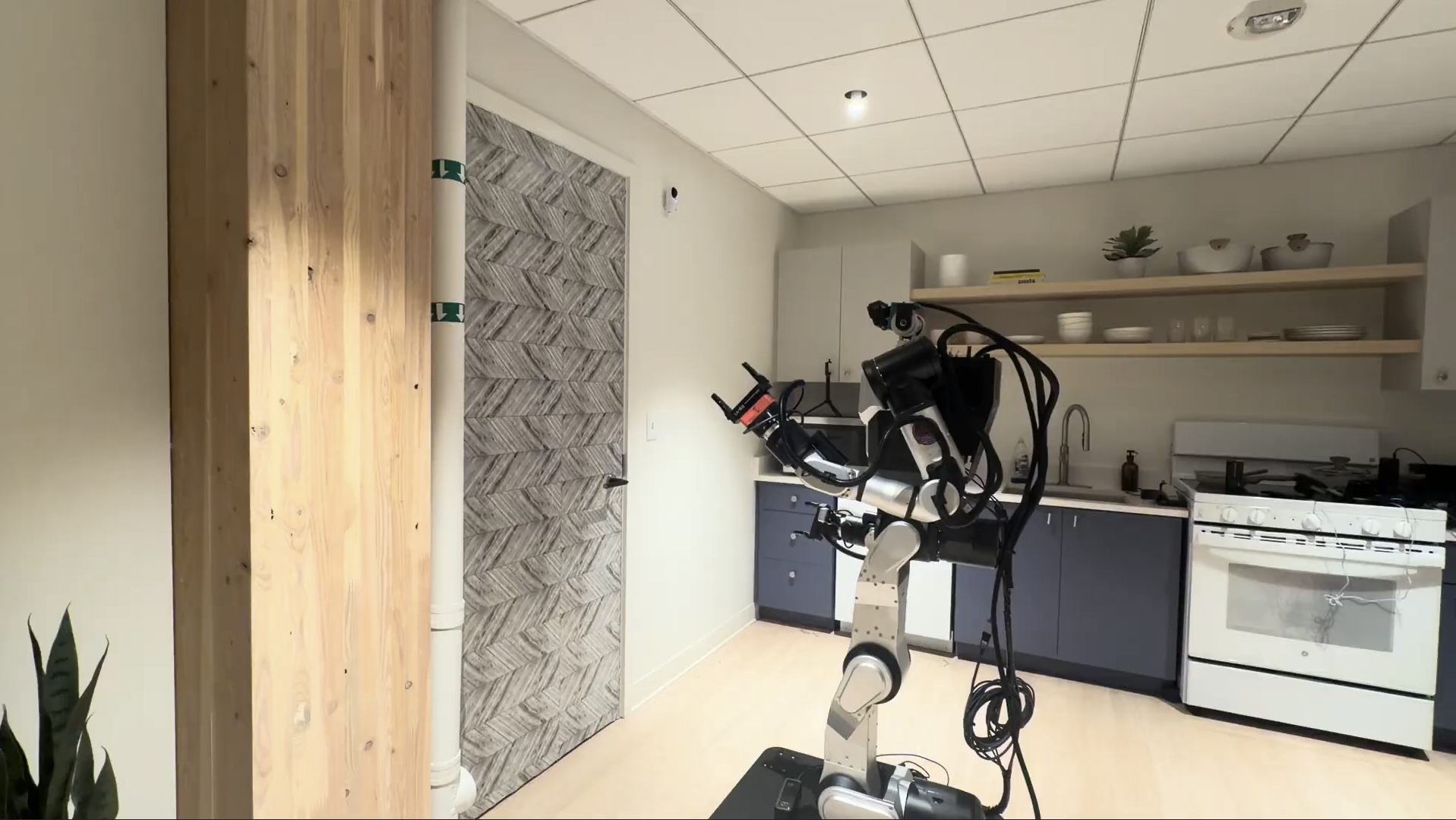}
    \includegraphics[width=0.24\linewidth,height=2.5cm,clip]{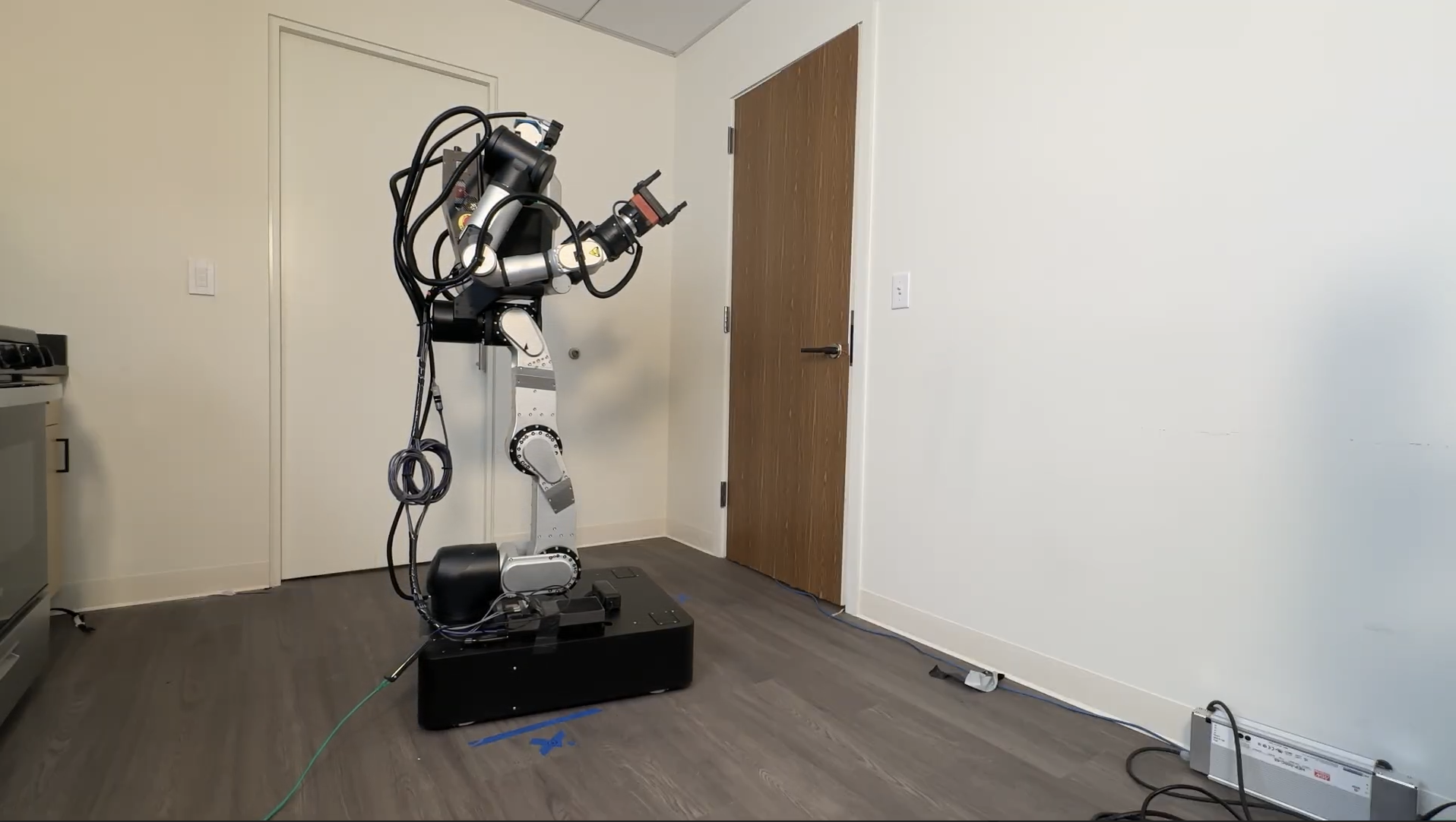}
    \includegraphics[width=0.24\linewidth,height=2.5cm,clip]{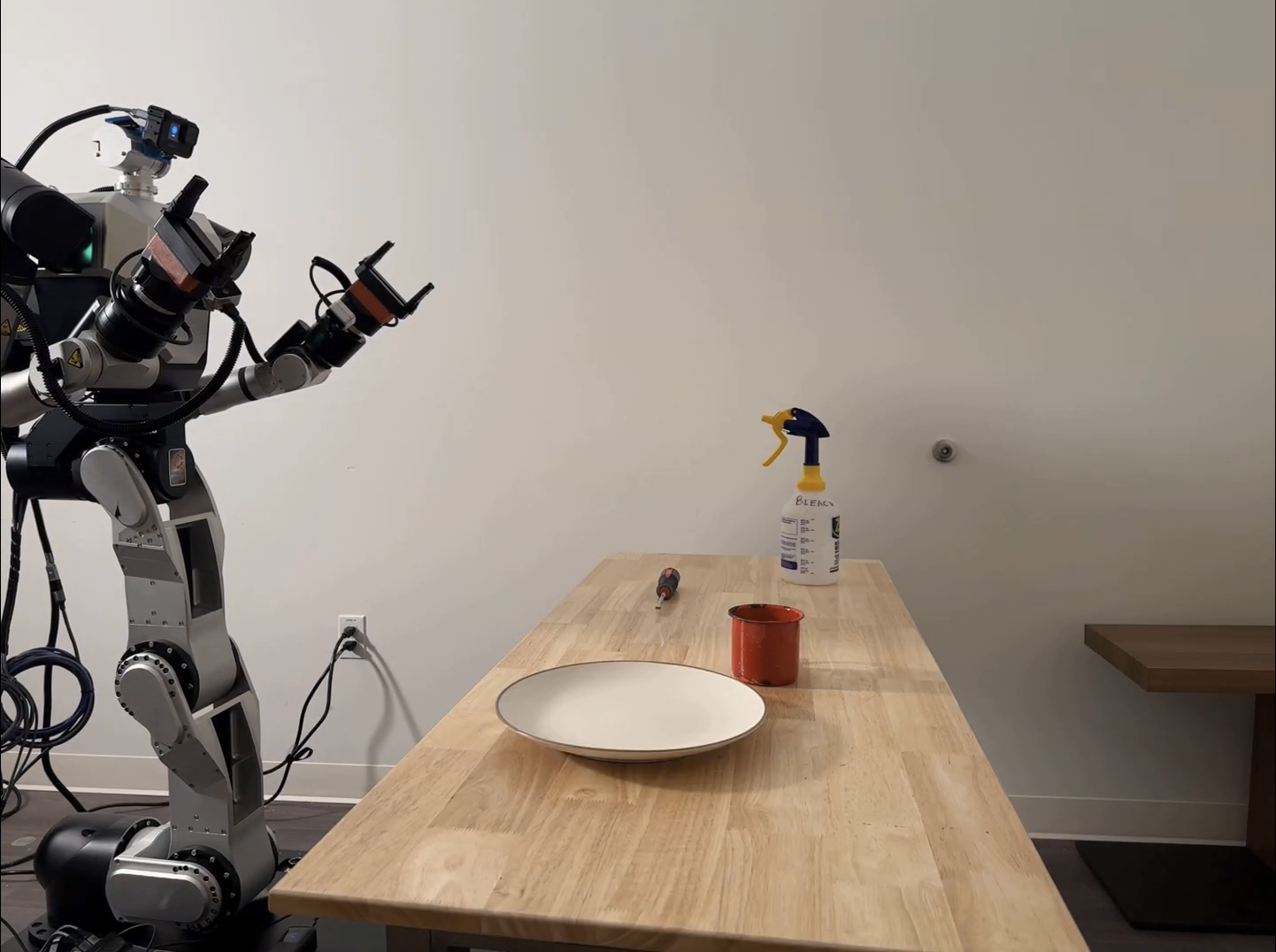}
    \caption{Real-world environments for our RBY1 articulated and rigid object mobile manipulation evaluations.}
    \label{fig:rby1_real_environments}
\end{figure*}

We evaluate the \molmomodel{} Door Specialist policy on a door opening task in three real-world environments, pictured in Fig.~\ref{fig:rby1_real_environments}, each featuring a distinct pull door with different visual textures, handle configurations, and surrounding scene context. Each environment also features distinct wall geometry and background clutter, testing the generalization of the policy across varied visual conditions. Episodes were executed at a 100ms inference timestep for safety reasons, which is slower than the timestep used during simulation evaluation.

\myparagraph{Failure analysis.} Aside from hardware failures, a recurring source of failure across Door 1 and Door 3 was difficulty grasping the handle. Both of these doors have handles positioned on the right side of the door, a configuration that is underrepresented in typical door interaction datasets and in our training data, which may explain the policy's reduced grasping reliability in these cases. In contrast, Door 2, whose handle configuration was more commonly represented, saw successful grasps in all three trials.

\begin{table}[t]
\centering
\caption{Door opening task results. Each door has a distinct visual texture.
Trials differ in robot base position. $^\dagger$Hardware fault occurred during trial.}
\label{tab:door_results}
\begin{tabular}{clccl}
\toprule
\textbf{Door} & \textbf{Trial} & \textbf{Grasp} & \textbf{Opened} & \textbf{Failure Mode} \\
\midrule
\multirow{3}{*}{1}
  & 1 & \cmark$^\dagger$ & \xmark & Grasping HW fault \\
  & 2 & \xmark           & \xmark & Base collision \\
  & 3 & \xmark           & \xmark & Joint limits \\
\midrule
\multirow{3}{*}{2}
  & 1 & \cmark$^\dagger$ & \xmark & Grasping HW fault \\
  & 2 & \cmark & \cmark$^\dagger$ & Opening HW fault \\
  & 3 & \cmark & \cmark$^\dagger$ & Opening HW fault \\
\midrule
\multirow{3}{*}{3}
  & 1 & \xmark           & \xmark & Base collision \\
  & 2 & \xmark$^\dagger$ & \xmark & Approach HW fault \\
  & 3 & \xmark           & \xmark & Grasping failure \\
\bottomrule
\end{tabular}
\end{table}

\subsection{Franka Simulation Evaluation}
\label{app:sim_eval}

\myparagraph{Setup.}
We run our experiments with the following hyper-parameters. A policy timestep of 66ms, task horizon of 303 steps / 20 seconds (pick tasks) and 606 steps / 40 seconds (for pick-and-place tasks). When using the filament renderer we set the environment illumination to 12000 candela by default. Additionally, we note that StereoVLA was trained based on a front-on view of the robot; in order to run StereoVLA we modify our benchmark by moving cameras into this position, and we additionally filter out episodes where the target object is not visible. This yields 92 episodes for the \textit{Pick v2} task and 91 episodes for the \textit{Pick v2 (F)} task. Despite these accommodations for StereoVLA, the performance remains low.

\myparagraph{Rendering.} While MuJoCo ships with a standard OpenGL renderer, MolmoSpaces~\cite{molmospaces2026} additionally provides a photorealistic material-based renderer using Filament. For some simulation evaluations, we use the Filament renderer for improved realism and further OOD evaluation settings.

\myparagraph{Tasks.}
Tab.~\ref{tab:sim_results} depicts our evaluation on a progression of increasingly difficult tasks. We begin with a simple pick task in a controlled configuration and limited object diversity (\textit{Pick v1}). The next set of tasks introduces more challenging object and viewpoint distributions in three variants: standard MuJoCo rendering (\textit{Pick v2}), photorealistic filament rendering (\textit{Pick v2 (F)}) which is out of distribution for our training data, and heavily randomized camera viewpoints (\textit{Pick v2 RC}). Pick tasks are allotted 20 seconds for completion. We then evaluate pick-and-place variants, including placing objects inside a receptacle (\textit{PnP Into}), next to a target (\textit{PnP Next-To}), and in a receptacle of a specified color (\textit{PnP Color}). We consider a Next-To episode as a success if the object to be moved is placed within $5$ cm of the target object. Pick-and-place tasks are allotted 40 seconds due to their increased difficulty and longer horizon. We report only oracle success (task completed at any timestep) for pick tasks, as termination behavior is not well-defined for object lifting. For pick-and-place tasks, we report both final success rate and oracle success rate. The gap between these metrics reflects whether the policy can recognize task completion and disengage appropriately rather than continuing to manipulate the object after placement.

\begin{table}[htb]
    \centering
    \addtolength{\tabcolsep}{-0.4em}
    \begin{tabular}{l l r l l}
        \toprule
        \textbf{Task Name} & \textbf{Objects} & \textbf{Episodes} & \textbf{Renderer} & \textbf{Camera} \\
        \midrule
        Pick v1      & Thor           & 1000   & MuJoCo   & DROID       \\
        Pick v2     & Objaverse & 200 & MuJoCo   & DROID-Light \\
        Pick v2 (F)            & Objaverse & 200 & Filament & DROID-Light \\
        Pick v2 RC (F)  & Objaverse & 200 & Filament & Random Camera   \\
        PnP-Next-To (F)     & Objaverse& 200 & Filament & DROID-Light \\
        PnP-Color       & Objaverse& 200 & MuJoCo & DROID-Light \\
        \bottomrule
    \end{tabular}
    \caption{Benchmark details for simulation tasks in Tab.~\ref{tab:sim_results}. ``Objects'' denotes the object library used to source the pickup object and placement receptacle. ``Camera'' describes the exocentric camera configuration: ``DROID'' is a fixed camera in an over-the-shoulder viewpoint (similar to the DROID dataset), ``DROID-Light'' is the same but with light random perturbation, and ``Random Camera'' is a completely random placement of the exo camera near the workspace (with guaranteed visibility of the task-relevant objects).}
    \label{tab:sim_task_benchmarks}
\end{table}

\myparagraph{Results.}
Our models outperform all baselines across tasks. On the least-variation \textit{Pick v1} task, \molmomodelmultiframe{1} achieves 93.5\% success compared to 48.0\% for the strongest baseline (\pizerofiveft{}). The gap widens on more challenging distributions: on \textit{Pick v2 RC}, our models achieve 40--66\% success while $\pi_{0.5}$ variants reach only 8--30\%. Other VLA baselines (StereoVLA, LAP-VLA, X-VLA) fail almost entirely on our evaluation suite, with success rates below 7\% on most tasks.
On pick-and-place tasks, which require both grasping and placement, \molmomodel{} variants achieve 63--67\% oracle success on \textit{PnP Into}. The closest baseline is \pizerofiveft{}, which achieves 43.5\% on this task.
% The gap between final and oracle success indicates that policies often complete the placement but subsequently disturb the object. 
% \pizerofiveft{} achieves 43.5\% oracle success on this task. Notably, our models generalize to compositional instructions (\textit{PnP Color}) where they must identify the correct receptacle by color, achieving 57--62\% final success.
Averaging across simulation tasks, \molmomodelmultiframe{1} achieves 64.1\% compared to 10.1\% for \pizerofive{} zero-shot. \paligemmamodel{}, which uses the $\pi_0$ architecture trained on our data, achieves 41.8\%---substantially higher than \pizerofive{} zero-shot and competitive with \pizerofiveft{} on pick tasks---demonstrating that much of the performance gain comes from \trainingdata{} rather than architectural differences.
On real-world evaluation, \molmomodelmultiframe{1} achieves 79.2\% success, compared to 31.3\% for \pizerofive{}. Interestingly, \pizerofiveft{} is competitive with our best policies on \textit{PnP Next-To}, achieving best oracle success though not final success. This supports the notion that fine-tuning real-world-native policies on sim data to bridge the real-to-sim gap enables reasonable comparisons, further bolstered by our results in the real world.

\subsection{RB-Y1 Simulation Evaluation}
\label{app:rby1_sim_eval}

Table \ref{tab:sim_rby1_results} reports zero-shot simulation performance across all RB-Y1 policies. \molmomodel{} Multitask outperforms \spocmodel{} across all shared tasks, which we attribute to several factors. First, \molmomodel{}'s frozen VLM backbone provides rich visual representations that generalize well without task-specific fine-tuning, whereas \spocmodel{}'s smaller transformer architecture has more limited capacity. Second, \molmomodel{} Multitask was trained jointly across all tasks, which may have enabled positive transfer between related manipulation behaviors. Although \spocmodel{} demonstrates more modest performance in these evaluations, its compact scale enables future on-policy reinforcement learning in simulation, which has been shown to yield substantial performance gains~\cite{hu2024flare}.

\subsection{Fixed-Camera Evaluation}

\begin{table}[t]
\newcommand{\ci}[1]{}  % <-- disable ci
\centering
\caption{Simulation and real evaluation with restricted camera setup. Success rate averaged over 1000 episodes in simulation and the 30 real-world Pick-and-Place trials from the Kitchen environment. All models evaluated zero-shot without task-specific fine-tuning.}
\label{tab:droid_sim_results}

\begin{tabular}{l|c|cc}
& & \multicolumn{2}{c}{\textbf{Real-World}} \\
\textbf{Model} & \textbf{Pick v1 (Sim)} & \textbf{Kitchen} & \textbf{All-Scene Avg} \\
\toprule
StereoVLA~\cite{deng2025stereovla} & 6.6\ci{1.7} & -- & -- \\
LAP-VLA~\cite{zha2026lap} & 19.4\ci{2.6} & -- & -- \\
\pizero~\cite{black2024pi_0} & 13.5\ci{2.3} & 20.0\ci{18.6} & 9.2 \\
\pizerofive~\cite{black2025pi_05} & 18.1\ci{2.5} & 63.3\ci{19.5} & 39.2 \\
\pizerofive-Finetune & 48.0\ci{3.2} & -- & -- \\
DreamZero~\cite{ye2026dreamzero} & 44.3\ci{3.1} & -- & -- \\
\midrule
\paligemmamodel{} & 66.2\ci{3.0} & 53.3\ci{19.0} & 46.7\\
\spocmodel{} & 70.4\ci{2.9} &  36.6\ci{19.5} & -- \\
\molmomodelimg{} & 92.2\ci{1.8} & 86.6\ci{17.4} & 72.5 \\
\molmomodelmultiframe{1} & \textbf{93.5}\ci{1.7} & 70.0\ci{19.4} & 79.2 \\
\molmomodelmultiframe{2} & 91.3\ci{1.9} & 73.3\ci{19.2} & 75.0 \\
\bottomrule
\end{tabular}
\end{table}

In Sec.~\ref{sec:real_eval} and Sec.~\ref{sec:sim_eval}, we evaluate Franka policies in both sim and real with diverse exocentric camera poses, which allows us to capture policy performance in a wide variety of deployment configurations. However, the DROID dataset does not provide very diverse exo camera poses, limiting the distribution to primarily left and right shoulder viewpoints. This means that other camera poses are out-of-distribution from the DROID dataset, and could degrade performance for policies trained or post-trained on DROID, which include many of our baselines.

To control for this phenomenon and provide a boost to our baselines, we filter our evaluations for those with a DROID-style shoulder camera configuration (and add in some additional baselines), and present them in Tab.~\ref{tab:droid_sim_results}. Specifically, our simulated \textit{Pick v1} task (with 1000 trials) uses a fixed shoulder camera, as well as our pick-and-place evaluations (with 30 trials) in the Kitchen environment (described in App.~\ref{app:real_eval_details}).

This setting biases the evaluation distribution towards the training distributions of our baselines. However, even when the evaluation is modified in the baselines' favor, \modelfamily{} policies still comfortably outperform all baselines in the fixed-camera setting. Interestingly, in our real-world evaluations, we also see that our baselines (\pizero{} and \pizerofive{}) receive a huge performance boost by limiting the randomization of the exo camera, but in comparison, \modelfamily{} policies experience a much smaller change with mixed results. This indicates that our baselines are much more sensitive to the camera pose than \modelfamily{} models, which exhibit significantly better robustness.

These results confirm two findings. First, \modelfamily{} policies are much more robust to camera pose. Second, even when this advantage is controlled for, \modelfamily{} models still significantly outperform our baselines. Evidently, \modelfamily{} policies' better performance over baselines is not solely from better camera pose robustness, but also the other forms of extreme diversity in \trainingdata{}.

\subsection{Data Ablations}
\label{sec:ablations_details}

In Sec.~\ref{sec:ablations}, we studied various axes of how the scale and diversity enabled by \trainingdataengine{} affect downstream performance in simulation and the real world. We present complete details of the methodology here.

\myparagraph{Modeling and Training.} For all data ablation experiments, we train the \molmomodelimg{} architecture from scratch for 24k train steps with a batch size of 512, while all other hyperparameters are kept the same. We generate bespoke datasets for each data point in the ablation results, as described in Sec.~\ref{sec:ablations}, focusing solely on the \textit{pick} task.

\myparagraph{Evaluation Details.} Unlike Sec.~\ref{sec:real_eval}, the data ablation real-world evaluations are on the \textit{pick} task. The evaluations are performed with the DROID platform in the workroom environment (pictured in Fig.~\ref{fig:droid_real_environments}), but with different objects: a roll of blue tape, mug, aerosol can, banana, pill bottle, and wooden spoon. Evaluations are done in three groups of three objects, with five trials for each object. For object class ablations, we adjust the object list slightly such that two out of the six objects are strictly in unseen classes for all top 100: apple and screwdriver replace the tape and wooden spoon. Success is judged by the evaluator when the object is fully off the table by approximately 2cm or more. Our simulation results are for \textit{Pick v2}, as detailed in Sec.~\ref{sec:sim_eval}. For each picking evaluation, the task prompt is formatted as: ``pick up the \texttt{<object>}''.

\subsection{Model Ablations}
\label{sec:model_ablations}

\begin{figure}[htb]
    \centering
    \includegraphics[width=0.7\linewidth]{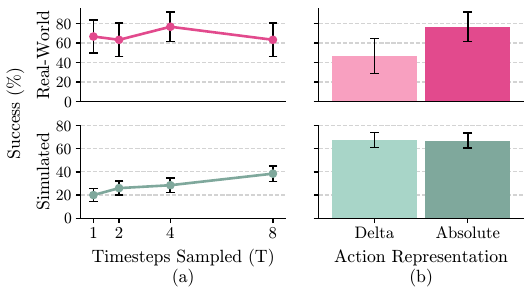}
    \caption{Ablations on training and action parameterization. We evaluate \molmomodelimg{} in the real world (top) and on \textit{Pick v2} in simulation (bottom). \textbf{(a)}~We train while sampling multiple denoising timesteps $T$ per example in parallel during training to improve convergence and final performance. We ablate $T \in \{1,2,4,8\}$ and find that simulation performance improves steadily as $T$ increases and peaks at $T=8$, while real-world performance is less monotonic and peaks at $T=4$. \textbf{(b)}~We train using either absolute or delta action representations for 200k steps on Franka FR3 policies. The absolute action representation substantially improves real-world performance over delta actions, while the two representations perform similarly in simulation. Error bars denote 95\% binomial proportion confidence intervals.}
    \label{fig:ablation_action}
\end{figure}

As in Sec.~\ref{sec:ablations}, our model ablation experiments are run in the real-world on the \textit{pick} task in the workroom environment, and on \textit{Pick v2} in simulation.

\myparagraph{Timesteps sampled during training} We sampled multiple time steps $T$ per example and denoise in parallel during training to improve the convergence and the accuracy of the model. We ablate the choice for $T \in \{1,2,4,8\}$ during training and report the performance of \molmomodelimg{} (Figure~\ref{fig:ablation_action}). Performance on simulation benchmarks improves as $T$ increases and peaks at $T=8$, suggesting the increase $T$ helps. However, while the result on real subset of 30 examples is not as clear, with the performance peaking at $T=4$.

\myparagraph{Action representation.} We compare \molmomodelimg{} trained using absolute and delta representations on the complete multi-task data mix for 200k steps each for the Franka FR3 policies (Figure~\ref{fig:ablation_action}). On the Franka FR3 task, the absolute policy significantly outperforms the delta policy in real setting, while the simulation results are on-par for both policies. The significant gap in the real world across 3 of our benchmarks strongly suggests that absolute joint policy models transfer better to real-world tasks.

\subsection{Compute-Controlled History-Conditioning Experiments}

In Sec.~\ref{sec:model_molmobot} we introduce \molmomodel{} with multiple variants, differing by the length of observation history. \molmomodelimg{} only takes the current frame, while \molmomodelmultiframe{1} and \molmomodelmultiframe{2} take 1 and 2 additional frames of history, respectively. As described in Sec.~\ref{sec:model_molmobot}, the history-conditioned variants are trained by initializing from \molmomodelimg{} and training for 50k additional timesteps with history. This means that \molmomodelimg{} has received 50k less train steps than \molmomodelmultiframe{1} and \molmomodelmultiframe{2}, so comparisons are not necessarily compute-controlled.

We address this by training \molmomodelimg{} for an additional 50k steps with the same learning rate schedule as the history-conditioned models to achieve compute parity, and name the resulting model ``\molmomodelimg{}+''. We evaluate this model in the workroom environment (Fig.~\ref{fig:droid_real_environments}) on pick-and-place for 30 total trials as described in Sec.~\ref{app:real_eval_details}, and report its performance along with other \molmomodel{} policies in Table~\ref{tab:molmobotimg_plus}. We see that some of the gains reported by history-conditioning may actually stem from additional training, but our evaluations cannot conclusively determine to what extent that may be the case. Our experiments show that the \molmomodel{} architecture is compatible and works with history conditioning, and we leave further exploration of how to best incorporate observation history to future work.

\begin{table}[htb]
    \centering
    \newcommand{\ci}[1]{{\color{gray}\scriptsize{$\,\pm$#1}}}
    \begin{tabular}{lc}
        \toprule
        Policy & Success Rate (\%) \\
        \midrule
        % MolmoBot-Pi0 & 60.0\ci{10.0} \\
        MolmoBot-Img & 76.7\ci{10.0} \\
        MolmoBot-Img+ & 83.3\ci{13.3} \\
        MolmoBot (H=1) & 90.0\ci{6.7} \\
        MolmoBot (H=2) & 76.7\ci{13.3} \\
        \bottomrule
    \end{tabular}
    \caption{MolmoBot policy real-world pick-and-place performance on 30 trials in the workroom environment, with 95\% CI half-width derived from stratified bootstrapping.}
    \label{tab:molmobotimg_plus}
\end{table}

\subsection{Zero-Shot Evaluation on External Simulation Benchmarks}
\label{sec:external_benchmarks}

Prior works often evaluate on popular existing benchmarks, including SIMPLER~\cite{li2024evaluating} and LIBERO~\cite{liu2023libero}. Both benchmarks were designed to evaluate policies trained on specific demonstration datasets (RT-1/Bridge and LIBERO demonstrations, respectively) and correlate with real-world performance under those in-distribution conditions. Additionally, these benchmarks use specific robot embodiments (WidowX and Franka FR3 with Panda Hand, respectively), which further complicates zero-shot evaluation of generalist policies. To evaluate on these benchmarks without benchmark-specific fine-tuning, a policy would have to be not only generalizable, but also cross-embodiment, posing a very high bar.

Therefore, we introduce and evaluate on SIMPLER-DROID and LIBERO-DROID. We replace the WidowX in SIMPLER and the non-DROID Franka in LIBERO with the DROID platform, which enables zero-shot evaluation of DROID policies, improving unification and reproducibility within the community. Specifically, we remove the existing robot from these benchmarks and replace it with a DROID platform, offset by a fixed transformation to account for differing embodiment size. Cameras are also swapped out, mirroring the DROID camera setup (ZED-Mini wrist camera and ZED 2 exocentric camera).

We evaluate \modelfamily{} policies and baselines---using no in-domain data or task-specific fine-tuning---on these adapted benchmarks, and our findings suggest structural limitations in their suitability for assessing generalist policies.

\myparagraph{SIMPLER.} SIMPLER provides simulated evaluation environments designed to correlate with real-world performance for policies trained on RT-1 and BridgeData V2. We reimplemented four WidowX tasks for our Franka setup. Results are shown in Table~\ref{tab:simpler_results}.

\begin{table*}[h]
\centering
\caption{Zero-shot evaluation on SIMPLER tasks reimplemented for DROID Franka FR3.}
\label{tab:simpler_results}
\small
\begin{tabular}{lcccc|c}
\toprule
\textbf{Model} & \textbf{Carrot$\rightarrow$Plate} & \textbf{Eggplant$\rightarrow$Basket} & \textbf{Stack Cubes} & \textbf{Spoon$\rightarrow$Cloth} & \textbf{Avg.} \\
\midrule
\molmomodelimg{} & 45.8\%  & 0\%  & 0\%  & 0\%  & 11.5\% \\
\paligemmamodel{} & 0\% & 0\%  & 0\%  & 0\%  & 0\% \\
\pizerofive{} & 0\%  & 0\%  & 0\%  & 0\%  & 0\% \\
\pizero{} & 0\%  & 0\% & 0\%  & 0\%  & 0\% \\
\bottomrule
\end{tabular}
\end{table*}

\molmomodelimg achieves 45.8\% on ``put carrot on plate,'' but all models fail on the remaining tasks. We found that the prompts used in the dataset were out of distribution for most models and need fine-tuning on that type of task. As reported, the task with the prompt ``put spoon on towel,'' led to 0\% performance for all models but tweaking it to ``put the spoon on the towel,'' led to more successful episodes. We leave prompt improvement to future work and report results with the exact prompt in the benchmark.

\myparagraph{LIBERO.} LIBERO is a lifelong learning benchmark with 130 manipulation tasks. We evaluate on LIBERO-Object, which requires picking specific grocery products (e.g., ``alphabet soup,'' ``bbq sauce'') and placing them in a basket. To diagnose failure sources, we evaluate both with and without distractor objects. Results are shown in Table~\ref{tab:libero_results}.

\begin{table*}[h]
\centering
\caption{Zero-shot evaluation on LIBERO-Object reimplemented for DROID Franka FR3. We evaluate with the standard set of distractors (std.) and without distractors (no dist.) to isolate failure sources.}
\label{tab:libero_results}
\small
\begin{tabular}{l|cc|cc|cc|cc}
\toprule
& \multicolumn{2}{c|}{\textbf{\molmomodelimg}} & \multicolumn{2}{c|}{\textbf{\paligemmamodel{}}} & \multicolumn{2}{c|}{\textbf{$\pi_{0.5}$-DROID}} & \multicolumn{2}{c}{\textbf{\pizero{}}} \\
\textbf{Task} & std. & no dist. & std. & no dist. & std. & no dist. & std. & no dist. \\
\midrule
Alphabet soup   & 0\%   & 16\%  & 0\% & 2\%  & 0\% & 2\% & 0\% & 0\% \\
Cream cheese    & 0\%   & 24\%  & 0\% & 2\%  & 2\% & 0\% & 0\% & 0\% \\
Salad dressing  & 20\%  & 40\%  & 0\% & 2\%  & 0\% & 0\% & 0\% & 0\% \\
BBQ sauce       & 0\%   & 42\%  & 0\% & 14\% & 0\% & 2\% & 0\% & 0\% \\
Ketchup         & 0\%   & 42\%  & 0\% & 6\%  & 0\% & 4\% & 0\% & 0\% \\
Tomato sauce    & 0\%   & 52\%  & 2\% & 6\%  & 2\% & 8\% & 0\% & 0\% \\
Butter          & 0\%   & 2\%   & 0\% & 0\%  & 0\% & 2\% & 0\% & 0\% \\
Milk            & 0\%   & 26\%  & 4\% & 2\%  & 4\% & 0\% & 0\% & 0\% \\
Choc.\ pudding  & 10\%  & 78\%  & 0\% & 2\%  & 0\% & 6\% & 0\% & 0\% \\
Orange juice    & 50\%  & 44\%  & 2\% & 2\%  & 2\% & 0\% & 0\% & 2\% \\
\midrule
\textbf{Average} & 8\% & 36.6\% & 0.8\% & 3.8\% & 1\% & 2.4\% & 0\% & 0.2\% \\
\bottomrule
\end{tabular}
\end{table*}

\molmomodelimg outperforms all baselines (36.6\% vs.\ 3.8\% average in the no-distractor setting), preserving the rank ordering observed in our real-world evaluations. The no-distractor ablation shows that visual clutter contributes to failure, but even simplified scenes yield low performance---the task descriptors reference specific brand-name products whose visual appearance differs from objects in MolmoBot-Data. The high task-level variance (78\% on ``chocolate pudding'' vs.\ 2\% on ``butter'') further suggests performance depends on incidental overlap with benchmark-specific assets rather than general manipulation competence.

\myparagraph{Discussion.} These benchmarks measure familiarity with specific assets and scene configurations rather than zero-shot manipulation capability. While \modelfamily{} models preserve rank-order superiority over baselines---consistent with our real-world findings---the absolute performance is not indicative of the models' true manipulation competence. In contrast, our MolmoSpaces-based evaluation (Sec.~\ref{sec:sim_eval}) uses 94k+ procedurally generated environments with 11k+ object assets, providing sufficient diversity to assess generalization while maintaining validated correlation with real-world performance.

% \subsection{Limitations and Future Work}
% \label{app:limitations}

% While this work represents a promising step for scaling up simulation-based pre-training, there are still several avenues of improvement for \molmomodel{}. \trainingdataengine{} is fundamentally constrained by assets that can currently be accurately simulated. We focus on rigid body and articulated object manipulation---tasks where modern simulators provide sufficient fidelity for transfer. Extending to contact-rich manipulation (e.g., insertion, peg-in-hole), deformable objects (cloth, rope, food), or tasks requiring accurate fluid or granular dynamics remains an open challenge. We believe that coupled with advances in physics-based and generative \emph{world model} simulators, our recipe of massive-scale procedural generation may extend to these more challenging tasks requiring contact-rich dexterity and deformables. We are excited by the potential frontiers of open research this work enables.

\end{document}